\documentclass{article}

\usepackage[T1]{fontenc}
\usepackage{microtype}
\usepackage{graphicx}
\usepackage{subcaption}
\usepackage{booktabs}
\usepackage{etoolbox}
\usepackage{caption} 
\usepackage{amsmath}
\usepackage{amssymb}
\usepackage{mathtools}
\usepackage{amsthm}
\usepackage{tikz}
\usetikzlibrary{trees, positioning}
\usepackage{multirow}
\usepackage{xcolor}
\usepackage{colortbl}
\usepackage{adjustbox}
\usepackage{listings}
\usepackage[most]{tcolorbox}
\usepackage{fvextra}
\usepackage{wrapfig}
\usepackage{array}
\usepackage{xspace}
\usepackage{upquote}
\usepackage{hyperref}
\usepackage{enumitem}

\usepackage[accepted]{icml2026}

\usepackage[capitalize,noabbrev]{cleveref}

\newcommand{\ours}{\textsc{DiscoverLLM}}

\definecolor{methods}{RGB}{225, 240, 255} 
\definecolor{improvement}{RGB}{255, 243, 224} 

\newcommand{\hlc}[2][yellow!30]{{\setlength{\fboxsep}{1pt}\colorbox{#1}{#2}}}

\newcommand{\teaser}{
  \centering
  \includegraphics[width=0.87\linewidth]{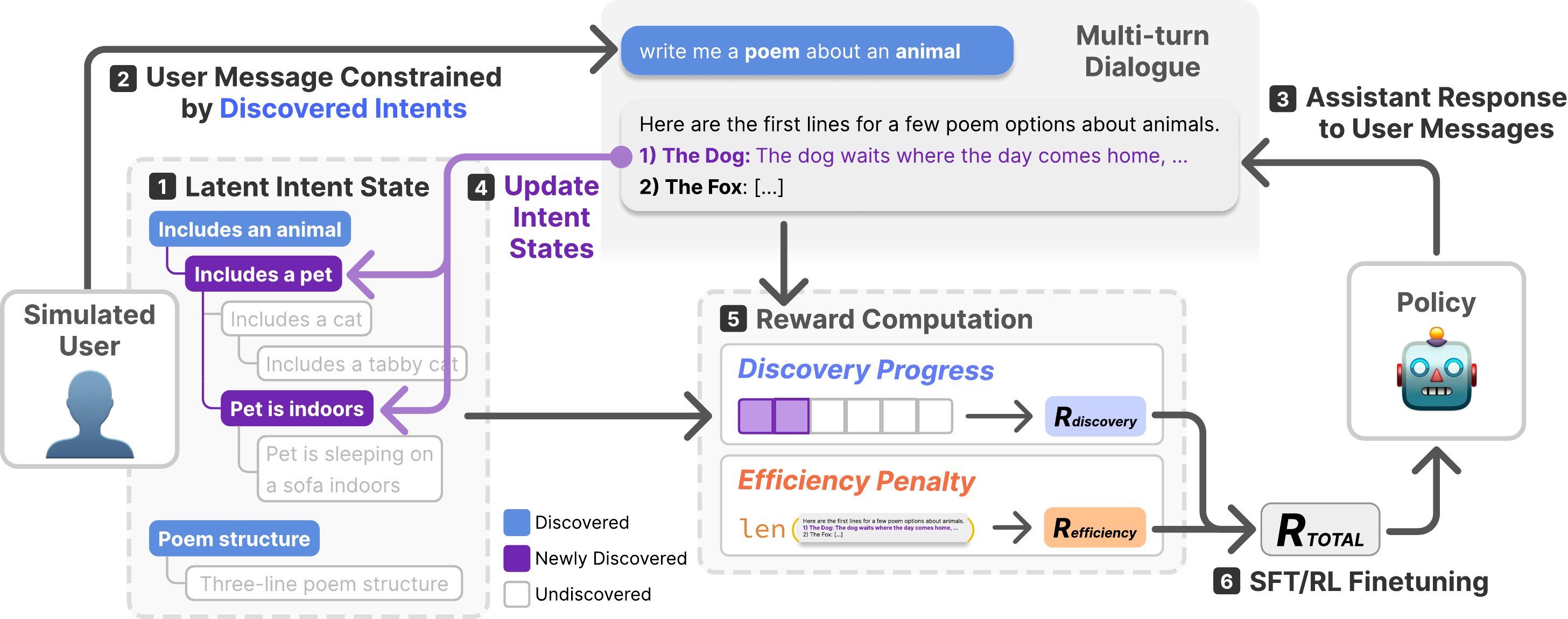}
  \captionof{figure}{\ours{} Framework: A simulated user with a latent intent hierarchy (1) interacts with a model. The user can only articulate discovered intents (2), and model responses (3) that successfully probe or satisfy undiscovered intents trigger state updates (4). The framework computes rewards based on discovery progress (5), which are used for fine-tuning of the model (6).}
  \label{fig:teaser}
}

\icmltitlerunning{\ours{}: From Executing Intents to Discovering Them}

\begin{document}

\twocolumn[
  \icmltitle{\ours{}: From Executing Intents to Discovering Them}



  \icmlsetsymbol{equal}{*}

  \begin{icmlauthorlist}
    \icmlauthor{Tae Soo Kim}{kaist}
    \icmlauthor{Yoonjoo Lee}{umich}
    \icmlauthor{Jaesang Yu}{kaist}
    \icmlauthor{John Joon Young Chung}{midjourney}
    \icmlauthor{Juho Kim}{kaist,skillbench}
  \end{icmlauthorlist}

  \icmlaffiliation{kaist}{KAIST}
  \icmlaffiliation{umich}{University of Michigan}
  \icmlaffiliation{midjourney}{Midjourney}
  \icmlaffiliation{skillbench}{SkillBench}
  
  \icmlcorrespondingauthor{Tae Soo Kim}{taesoo.kim@kaist.ac.kr}

  \icmlkeywords{Large Language Models, Human-Centric AI, User Simulator}

  \begin{center}            \href{https://taesookim.com/discoverllm}{\texttt{taesookim.com/discoverllm}}
  \end{center}

  \vskip 0.15in
  \teaser
  \vskip 0.15in
]



\printAffiliationsAndNotice{}  % no special notice (required even if empty)

\begin{abstract}
\vspace{-4pt}
To handle ambiguous and open-ended requests, Large Language Models (LLMs) are increasingly trained to interact with users to surface intents they have not yet expressed (e.g., ask clarification questions).
However, users are often ambiguous because they have not yet \textit{formed} their intents: they must observe and explore outcomes to discover what they want. 
Simply asking "what kind of tone do you want?" fails when users themselves do not know. 
We introduce \ours{}, a novel and generalizable framework that trains LLMs to help users form and discover their intents. 
Central to our approach is a novel user simulator that models cognitive state with a hierarchy of intents that progressively concretize as the model surfaces relevant options---where the degree of concretization serves as a reward signal that models can be trained to optimize.
Resulting models learn to collaborate with users by adaptively \textit{diverging} (i.e., explore options) when intents are unclear, and \textit{converging} (i.e., refine and implement) when intents concretize. 
Across proposed interactive benchmarks in creative writing, technical writing, and SVG drawing, \ours{} achieves over 10\% higher task performance while reducing conversation length by up to 40\%. 
In a user study with 75 human participants, \ours{} improved conversation satisfaction and efficiency compared to baselines.
\end{abstract}
\vspace{-20pt}
\section{Introduction}

Large Language Models (LLMs) have emerged as effective conversational assistants, capable of following complex user requirements to generate fluent and high-quality outputs.
However, this effectiveness relies on a critical assumption: users begin with fully formed intents.
In reality, users often approach tasks with \textit{ill-defined intents}, not knowing precisely what they want~\cite{subramonyam2024bridging}---which is common in diverse open-ended or creative tasks like writing and design~\cite{schon2017reflective, dow2010parallel, dorst2001creativity, flower1981cognitive}.
Consider a user who asks an LLM to ``write a personal essay about a hard lesson.''
The model produces a full essay with a neutral formal tone. 
After reading a few lines, the user feels it is not right but cannot pinpoint why so they ask ``maybe a unique tone?'' 
Noting the ambiguity, the assistant tries to clarify: ``What type of unique tone do you want?'' 
The user is unsure so they hesitantly say "not sure, something different" to which the assistant generates a highly personal and emotional essay, which feels like ``too much.'' 
A revision tones down the intimacy and emotion, making the user feel that it is ``too distant'' now.
Only after several more turns does the user realize what they want: a narrative that is emotionally restrained but still personally revealing.

Current approaches to improving LLM interaction do not address this challenge of \textit{ill-defined, unformed intents}.
Benchmarks mainly assess single-turn performance~\cite{laban2025llms} and fine-tuning techniques (e.g., RLHF~\cite{ouyang2022training}) reward single-turn full outputs---assuming all crucial details are present in the users' initial requests.
Recent work on improving models' multi-turn capabilities, whether through prompting~\cite{li2023eliciting, mu2023clarifygpt} or fine-tuning~\cite{zhang2024modeling, shani2024multi, wu2025collabllm}, assumes users possess well-defined intents that they have simply not articulated, which the model can surface by asking clarifying questions.
But when intents are not yet discovered or formed, there is nothing to surface---like how the example user could not answer what tone they wanted.

Research on the cognitive process of tackling open-ended, ill-defined problems offers an alternative: people's understanding of a problem and its solutions \textit{co-evolve}---people discover what they need (\textit{problem space}) by exploring possible outcomes (\textit{solution space})~\cite{cross1982designerly, dorst2001creativity, schon2017reflective}.
By creating and examining options, even incomplete ones, people gradually discover what they need or want---like the example user discovering their desired tone by seeing opposite options.
This process of \textbf{discovering intent through exploration} demands a fundamentally different role for LLMs: instead of simply eliciting and executing users' intents, models should help users explore, discover, and form their intents.

Inspired by this, we formalize \textit{intent discovery} as a distinct problem from intent elicitation and propose a novel \textit{user simulator} that operationalizes this formalization. 
Unlike prior simulators where users have hidden but fully formed goals~\cite{wu2025collabllm, sun2025training}, ours represent goals as a hierarchy of intents---from broad to specific---that progressively concretize when a model successfully satisfies or directly asks about them.
The simulator can reward model responses based on how effectively they help uncover more specific intents, enabling fine-tuning by synthesizing high-quality conversations or Reinforcement Learning (RL)~\cite{schulman2017proximal, shao2024deepseekmath}.
With this, we propose a training framework, \ours{}, that teaches language models to help users \textit{discover} and \textit{refine} their intents.
Returning to our example: a simulated user possesses the hierarchy of \textit{\textbf{unique tone} $\rightarrow$ personal but composed $\rightarrow$ personally revealing but emotionally restrained}. 
When only the top-level intent is formed, asking ``what type of unique tone?'' fails, but generating an overly personal draft leads the user to discover ``personal tone.''

Using our framework, we propose three challenging multi-turn tasks across diverse open-ended, creation domains: creative writing, technical writing, and SVG drawing.
On these tasks, we apply \ours{} to fine-tune Llama-3.1-8B-Instruct and Qwen3-8B. 
Across models and tasks, our approach improved intent discovery by around 10\%, increased interactivity scores by 83\% (as rated by LLM judges), and reduced conversation length by 32\%---all compared to the best baselines. 
We also verified that these gains generalized to unseen domains (e.g., travel planning, web development). 
In a user study with 75 crowdworkers, who completed writing tasks with anonymized models, \ours{} achieved higher user satisfaction while reducing task completion time---with participants noting how the model appeared to \textit{``anticipate''} their latent intents.
\section{Problem Formulation}

We consider multi-turn conversations where a user collaborates with an AI assistant to perform an open-ended task. 
Instead of assuming that users start with fully-formed intents, we formalize a setting where the user's \textit{intents are discovered through the conversation}.

\subsection{Intent Discovery Through Interaction}
\label{sec:problem_formulation}

At turn $t$ of a conversation $\mathcal{C} = (u_1, r_1, u_2, r_2, \ldots, u_T, r_T)$, the user sends message $u_t$ and receives assistant response $r_t$.
Let $I_t$ denote the user's intent state at turn $t$: a collection of requirements and constraints that must be satisfied in this task.
We model intent formation as progressive refinement:
\begin{equation}
    I_0 \subseteq I_1 \subseteq \cdots \subseteq I_T
\end{equation}
In a successful conversation, the initial state $I_0$ contains a few abstract requirements (e.g., \textit{``a poem''}, \textit{``includes an animal''}) while the final state $I_T$ has multiple concrete ones (e.g., \textit{``a haiku''}, \textit{``includes a tabby cat''})---each transition adding or concretizing requirements.

\paragraph{Response-Driven Progressive Discovery.} 
Drawing on cognition research that describes how people develop understanding of their goals by exploring or creating outcomes~\cite{schon2017reflective, flower1981cognitive}, we formalize intent discovery as dependent on the assistant's responses.

Let $\mathcal{R}(I_t)$ denote the set of \emph{potential refinements}: directions to concretize or expand $I_t$ that the user has not yet formed but could adopt if surfaced in the interaction.
For instance, if $I_t$ includes ``includes a pet,'' then $\mathcal{R}(I_t)$ might contain ``includes a cat'' or ``pet in an indoor setting.''

Let $\phi: (r_t, I') \to \{0, 1\}$ indicate whether the assistant's response $r_t$ successfully \textit{engages} with refinement $I' \in \mathcal{R}(I_t)$: directly asks about that aspect or reveals it through a concrete artifact that satisfies it.
The intent state updates as:
\begin{equation}
    I_{t+1} = I_t \cup \{I' \in \mathcal{R}(I_t) : \phi(r_t, I') = 1\}
\end{equation}
When a response engages refinements in $\mathcal{R}(I_t)$, the user may recognize them as matching latent preferences.  
For example, while initially only knowing they want a poem featuring \textit{``an animal''}, a response featuring \textit{``a dog''} leads the user to realize they want \textit{``a pet''}---adding this specification to $I_{t+1}$. 
A response that fails to engage any refinement leaves the intent state unchanged.

\paragraph{User Expressiveness Constraint.}
In $u_t$, users can \textit{only} articulate intents in $I_t$ and cannot request undiscovered refinements in $\mathcal{R}(I_t)$. 
A user who knows \textit{``pet''} but not the refinement \textit{``cat''} can only vaguely request ``maybe another type of pet?''
This asymmetry distinguishes our setting from standard intent elicitation: a clarifying question like ``what type of pet?'' fails to progress the conversation as the user has not realized the refinement and cannot yet answer.

\subsection{Objectives}

The assistant must balance two objectives: 
\textbf{(1) Intent Discovery:} Help the user discover concrete, specific intents by exploring and surfacing possibilities that engage with latent, undiscovered intents in $\mathcal{R}(I_t)$. 
\textbf{(2) Intent Satisfaction:} Produce outcomes that satisfy the user's discovered intents $I_T$, tracking and satisfying increasingly detailed requirements as they emerge. 
Effective assistance requires balancing \textit{divergence} (i.e., explore options to probe around abstract intents) and \textit{convergence} (i.e., refine outputs to integrate and satisfy concrete intents)~\cite{goldschmidt2016linkographic}.

\subsection{Underlying Assumptions}

Our formalization assumes users possess latent but unformed intents---a general direction that concretizes only once the assistant surfaces relevant options. 
In practice, users may lack any initial direction and construct intents entirely through interaction, giving the assistant greater influence over what intents form. 
Our formalization partially emulates this: since users cannot articulate intents unless surfaced by the assistant, they appear to construct intents from what is mentioned. 
However, as we model discovery rather than construction, the assistant can only surface possibilities users are already disposed toward, not direct them to new ones. 
We also assume monotonic refinement: once discovered, intents remain discovered---ignoring cases where users abandon or backtrack on intents.
Despite these assumptions, our formalization captures common patterns of intent concretization~\cite{flower1981cognitive, schon2017reflective}---relaxing these assumptions remains future work.
\section{\ours{}: General Training Framework for Intent Discovery}

We propose \ours{}, a training framework that operationalizes the intent discovery formulation.
The central challenge is obtaining a training signal: in natural conversations, we cannot observe a user's intent state $I_t$ or refinement space $\mathcal{R}(I_t)$. 
Our solution is a user simulator with an explicit latent intent structure, enabling reward computation for training.
Details and prompts in Appendix~\ref{appendix:user_simulator}.

\begin{figure*}[ht]
  \centering
  \includegraphics[width=0.92\linewidth]{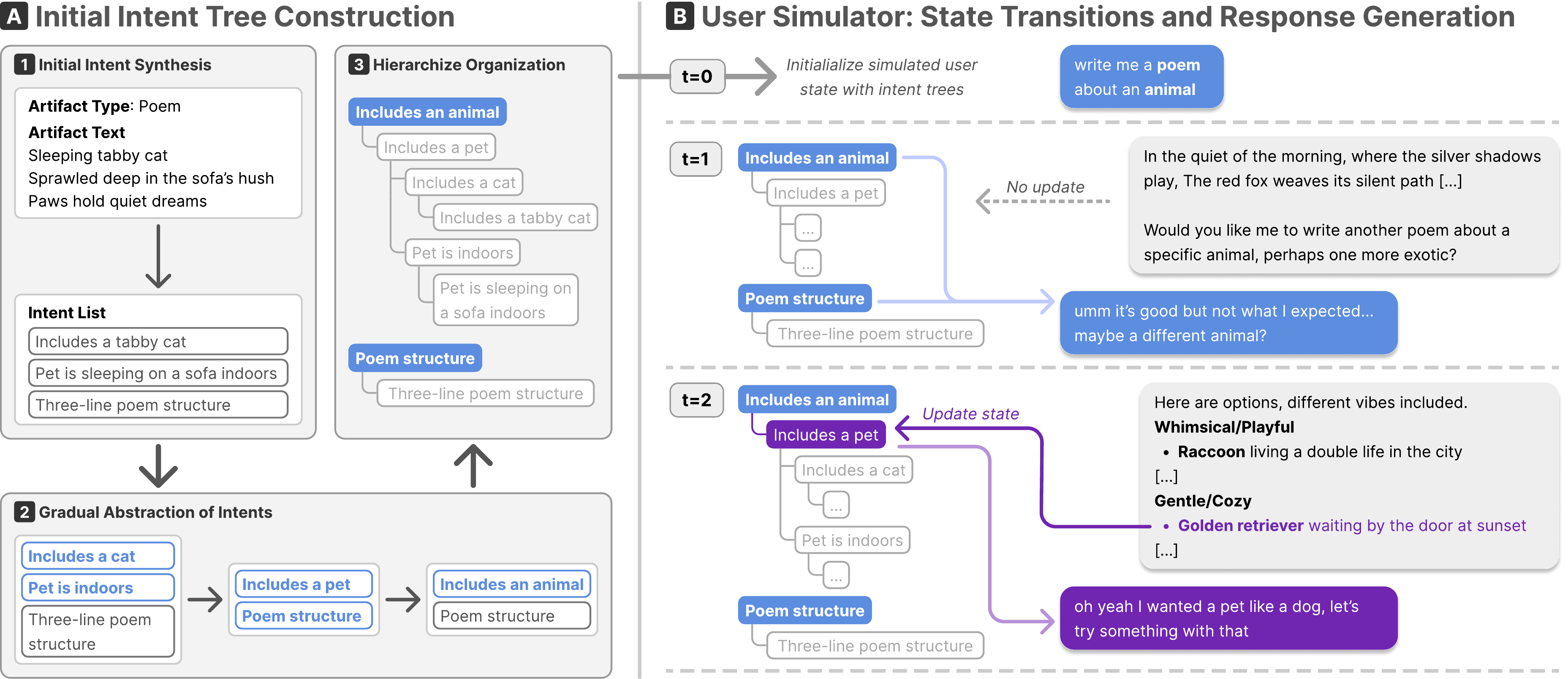}
  \caption{\textbf{(A) Intent tree construction}: (1) Given an artifact type and its content, we generate a specific intents list, (2) iteratively abstract them across levels, and (3) organize all resulting intents into a tree hierarchy. \textbf{(B) Simulation Example:} The user simulator begins ($t=0$) with only a few abstract intents discovered and provides an initial request based on these. At $t=1$, the model's response fails to probe or satisfy intents in the refinement space so no state updates occur and the user remains vague. At $t=2$, the model provides various options where one probes at an undiscovered intent, updating its state and enabling the user to articulate it.}
  \vspace{-8pt}
  \label{fig:user_simulator}
\end{figure*}

\subsection{Operationalizing Intents as a Hierarchy}

We represent intents as a tree hierarchy $\mathcal{H} = (V, E)$, where each node $v \in V$ is a requirement, and edges connect parents to more specific children.
This hierarchical structure is grounded in cognitive process theory that posits that people \textit{``create a hierarchical network of goals''}~\cite{flower1981cognitive}.
The hierarchy serves as ground truth for our simulator: it defines what the user would recognize as matching their preferences \emph{if} surfaced by the assistant, enabling reward computation for intent discovery.
A simulated user may have multiple trees $\{\mathcal{H}_1, \ldots, \mathcal{H}_K\}$ for different intent dimensions (e.g., subject, tone, structure). 
For simplicity, we use $V$ to denote all nodes across these trees.

The user's intent state $I_t \subseteq V$ is the set of discovered nodes by turn $t$, and the refinement space consists of undiscovered children of discovered nodes:
\begin{equation}
    \mathcal{R}(I_t) = \{v \in V : \text{parent}(v) \in I_t, \, v \notin I_t\}
\end{equation}
This captures \textbf{progressive discovery} (Sec.~\ref{sec:problem_formulation}): a node can only be discovered once its parent is discovered.
Branches are independent, so discovering intents along one path does not require resolving others.
In Figure~\ref{fig:user_simulator}, the user begins with \textit{``includes an animal''} from which \textit{``includes a pet''} can be discovered, and discovering ``pet'' enables discovery of the pet type (\textit{``cat''}) and setting (\textit{``indoors''}).

\paragraph{Discovery States.} 
Each node maintains a discovery state: \textit{undiscovered}, \textit{emerging}, or \textit{discovered}. 
Emerging nodes represent intents the user has begun to recognize but not fully realized---they may only be vaguely referenced (e.g., \textit{``maybe a smaller animal?''} for an emerging ``pet'' intent).
Only discovered nodes can be directly articulated, satisfying the \textbf{expressiveness constraint} (Sec.~\ref{sec:problem_formulation}).

\subsection{Response Evaluation \& State Transitions (Fig.~\ref{fig:user_simulator}B)}
\label{sec:framework_evaluation}

At each turn, the simulator evaluates how the response $r_t$ engages with nodes in $\mathcal{R}(I_t)$. 
If $r_t$ is an \textit{artifact} (e.g., draft, samples), whether it satisfies the intent.
If it is a \textit{dialogue act} (e.g., question, suggestion), whether it probes the intent.

\paragraph{State Transitions.} We model transitions to capture how both direct and indirect exposure can trigger intent discovery.
\textbf{Direct engagement:} If $r_t$ explicitly asks about or satisfies the intent, the node becomes \textit{discovered}.
\textbf{Tangential engagement:} If $r_t$ provides related but non-matching options, each contributes toward a cumulative score, where exceeding a randomly sampled threshold advances the node one state (\textit{undiscovered} $\to$ \textit{emerging} $\to$ \textit{discovered}).
This reflects how people discover preferences not only through exact matches, but by narrowing possibilities and seeing what they \textit{do not} want~\cite{schon2017reflective, slovic1995construction}.
Conditioned on the resulting discovery state, the simulator's generated responses naturally produces a full range of dialogue behaviors (e.g., feedback, refinement or execution requests, acceptance) without further explicit prompting.

\subsection{Constructing Intent Trees (\autoref{fig:user_simulator}A)}
\label{sec:framework_abstraction}

We construct intent trees automatically from existing artifacts (e.g., stories, code, etc.), enabling generalization across tasks and domains.
\textbf{(1) Initial Intent Synthesis:} Given artifact $a$, we prompt an LLM to list all requirements satisfied by the artifact, which become the most concrete intents of the hierarchy.
\textbf{(2) Gradual Abstraction:} An LLM iteratively abstracts or generalizes the intent list through multiple levels by abstracting intents or removing them at each level.
\textbf{(3) Hierarchy Organization:} Given the intents at multiple abstraction levels, an LLM organizes them into a tree by identifying abstract intents that subsume concrete ones.

\subsection{Reward Function}

We design the per-turn reward: $\text{R}(r_t) = \text{R}_{\text{d}}(r_t) + \text{R}_{\text{e}}(r_t)$.

\paragraph{Discovery Progress ($\text{R}_{\text{d}}$).} 
This reward measures progress toward complete intent specification: $\text{R}_{\text{d}}(r_t) = |I_{t+1}| - |I_t|$.
This captures both discovery and satisfaction: satisfying an intent also discovers it, and an intent must be discovered to be expressed and subsequently satisfied.

\paragraph{Efficiency Penalty ($\text{R}_{\text{e}}$).} To encourage concise interactions, we penalize excessive response length: $\text{R}_{\text{e}}(r_t) = -\min(\lambda \cdot \max(0, \text{tokens}(r_t) - \tau), 1)$, where $\tau$ is a token threshold and $\lambda$ controls penalty severity.

\paragraph{Design Rationale.} Our reward design prioritizes discovery first, then efficiency: each discovered intent contributes +1 while efficiency penalty is capped at 1. 
We avoided normalizing $\text{R}_{\text{d}}$ by remaining intents, as this yielded inconsistent signals: length penalties negated discovery gains, and later turns became unstable as the denominator shrunk.

\subsection{Training}

With our user simulator and the reward function, our framework supports multiple training paradigms.
We can simulate interactions between the user simulator and assistant models, sampling multiple responses per turn and selecting the top-ranked ones via the reward function.
This yields a high-quality dataset of synthetic conversations for \textbf{Supervised Fine-Tuning (SFT)}.
This can also yield pairwise comparisons at each conversation turn, which can be used for \textbf{Offline Reinforcement Learning (RL)}, like DPO~\cite{rafailov2023direct}.
Finally, models can be optimized directly on the reward signal through \textbf{Online RL} methods, like PPO~\cite{schulman2017proximal} or GRPO~\cite{shao2024deepseekmath}.

\section{Experimental Setup}

We apply the \ours{} framework to create multi-turn datasets for both fine-tuning and evaluation across three diverse domains: creative writing, technical writing, and visual design. 
Details in Appendix~\ref{appendix:evaluation_details}.

\subsection{Tasks and Datasets}

We focus on open-ended tasks that require substantial user-assistant collaboration and iteration. 
Writing---among the most common uses of AI assistants~\citep{tamkin2024clio, zao2024people}---involves iterative composition and revision~\citep{flower1981cognitive}. 
Visual creation tasks like SVG drawing share similar properties: users often concretize preferences through exploration~\citep{lee2010designing}.

We construct datasets from existing artifact sources: \textbf{Creative Writing}~\citep{fan2018hierarchical}\footnote{We used the dataset in \url{https://huggingface.co/datasets/euclaise/WritingPrompts_preferences}} (i.e., posts in the r/WritingPrompts subreddit), \textbf{Technical Writing}~\citep{roberts2021media} (i.e., journalistic articles), and \textbf{SVG Drawing}~\citep{xing2024llm4svg}. 
From each, we sample 500 artifacts for training and 100 for evaluation. 
We use Claude Sonnet 4.5 to construct intent trees (Sec.~\ref{sec:framework_abstraction}) and Gemini 3 Flash for the user simulator (i.e., state transition and response generation in Sec.~\ref{sec:framework_evaluation}). 
Each conversation runs 5 turns and, during evaluation, we repeat each conversation 3 times and average results.
Further details in Appendix~\ref{appendix:dataset_generation}.

\vspace{-2pt}
\subsection{Evaluation Metrics}

We simulate conversations between each user simulator and evaluated model to assess performance on four metrics.

\vspace{-4pt}
\paragraph{Intent Discovery Score.} 
We measure the proportion of intents discovered by the end of each conversation. 
To account for variation in difficulty across artifacts, we normalize scores using per-instance bounds: exclude intents discovered by \emph{all} evaluated models (i.e., trivially easy) and those discovered by \emph{none} (i.e., unreachable within the conversation length).
This focuses evaluation on the discriminative range where model behavior meaningfully differs:
\vspace{-6pt}
\begin{equation}
    \textit{Discovery} = \frac{1}{N} \sum_{i=1}^{N} \frac{|I_T^{(i)}| - |I_{\text{all}}^{(i)}|}{|I_{\text{any}}^{(i)}| - |I_{\text{all}}^{(i)}|}
\end{equation}
\vspace{-2pt}
where $I_T^{(i)}$ is the set of discovered intents at conversation end for artifact $i$, $I_{\text{all}}^{(i)}$ is the set discovered by all models, and $I_{\text{any}}^{(i)}$ is the set discovered by at least one model.

\vspace{-4pt}
\paragraph{Intent Satisfaction Score.} 
To assess how intent discovery leads to better intent satisfaction, we append a message at the end of each conversation prompting the model to generate a final complete artifact.
We then use an LLM-as-a-judge~\citep{zheng2023judging} (GPT-5.1) to assess whether the artifact satisfies each leaf node in the intent trees (e.g., most specific intents from the original artifact).
The satisfaction score is the proportion of satisfied leaf nodes, but excluding those that were not satisfied by any evaluated model.

\vspace{-4pt}
\paragraph{Interactivity Score.} Following \citet{wu2025collabllm}, we use an LLM judge (GPT-5.1) to rate how well the assistant collaborates and engages with the user in each conversation---scores rescaled to 0-1.

\vspace{-4pt}
\paragraph{Average Token Count.} We compute the mean number of tokens generated by each model across all turns.

\subsection{Training \ours{}s}

We apply the \ours{} framework to two base models: Llama-3.1-8B-Instruct~\citep{grattafiori2024llama} and Qwen3-8B~\citep{yang2025qwen3}, using LoRA fine-tuning~\citep{hu2021lora}. 
We train four model variants with progressively more optimization:
\textbf{(1) SFT}: Supervised fine-tuning on synthesized conversation histories. 
\textbf{(2) DPO}: Starting from the base model, we apply Offline DPO on preference pairs from the same synthesized conversations.
\textbf{(3) SFT+DPO}: We apply Offline DPO but starting with the SFT model.
\textbf{(4) SFT+DPO+GRPO}: For the Qwen3-8B-based models, we additionally apply Online GRPO.
Details in \ref{appendix:training_details}.

\subsection{Baselines}

We compare \ours{} models against three baselines: (1) Llama-3.1-8B-Instruct and Qwen3-8B (\textit{Base}), (2) the base models with the same system prompt (Appendix \ref{prompt:assistant_evaluation}) that instructs them to support the user's intent discovery (\textit{Prompted Base}), and (3) \textsc{CollabLLM}~\citep{wu2025collabllm}, a fine-tuning of Llama-3.1-8B-Instruct trained to collaborate with users through follow-ups and clarifications.
\begin{table*}[ht]
\centering
\begin{adjustbox}{max width=\textwidth}
\small
\begin{tabular}{l cccc cccc cccc}
\toprule
& \multicolumn{4}{c}{\textbf{Creative Writing}} & \multicolumn{4}{c}{\textbf{Technical Writing}} & \multicolumn{4}{c}{\textbf{SVG Drawing}} \\
\cmidrule(lr){2-5} \cmidrule(lr){6-9} \cmidrule(lr){10-13}
& \textbf{Discover}$\uparrow$ & \textbf{Satisfy}$\uparrow$ & \textbf{ITR}$\uparrow$ & \textbf{\#Tok$(k)$}$\downarrow$ & \textbf{Discover}$\uparrow$ & \textbf{Satisfy}$\uparrow$ & \textbf{ITR}$\uparrow$ & \textbf{\#Tok$(k)$}$\downarrow$ & \textbf{Discover}$\uparrow$ & \textbf{Satisfy}$\uparrow$ & \textbf{ITR}$\uparrow$ & \textbf{\#Tok$(k)$}$\downarrow$ \\
\midrule
\multicolumn{13}{l}{\textbf{Llama-3.1-8B-Instruct}} \\
\quad Base 
    & 38.2 & 30.0 & 20.1 & 3.09 
    & 49.1 & 36.0 & 21.2 & 3.32 
    & 45.6 & 32.5 & 21.6 & 3.59 \\
\quad Prompted Base 
    & 37.7 & 26.4 & 26.0 & 2.97 
    & 43.6 & 33.5 & 24.2 & 3.05 
    & 40.0 & 30.9 & 25.1 & 3.18 \\
\quad \textsc{CollabLLM} 
    & 37.3 & 28.0 & 32.6 & 2.93 
    & 45.8 & 33.7 & 24.9 & 3.13
    & 43.0 & 29.9 & 30.8 & 3.18 \\
\rowcolor{methods}
\quad SFT 
    & 40.7 & 33.4 & 92.3 & 1.71 
    & 47.1 & 35.2 & 81.6 & 2.09 
    & 45.4 & 34.9 & 66.9 & 2.92 \\
\rowcolor{methods}
\quad DPO 
    & 40.5 & 29.2 & 33.1 & 2.91 
    & 47.2 & 34.2 & 27.3 & 3.11
    & 45.3 & 32.5 & 29.2 & 2.89 \\
\rowcolor{methods}
\quad SFT+DPO 
    & 42.4 & 28.4 & 32.9 & 2.77 
    & 49.0 & 35.9 & 31.3 & 2.94
    & 51.6 & 37.0 & 44.6 & 2.61 \\
\rowcolor{improvement}
\quad Rel.\ Improv. 
    & 11.0\% & 11.3\% & 183\% & 44.7\% 
    & -0.0\% & -0.0\% & 227\% & 11.4\% 
    & 13.2\% & 13.8\% & 117\% & 27.3\% \\

\midrule

\multicolumn{13}{l}{\textbf{Qwen3-8B}} \\
\quad Base 
    & 35.2 & 30.4 & 36.2 & 3.41 
    & 40.7 & 33.7 & 35.3 & 3.39 
    & 47.0 & 32.0 & 54.4 & 3.96 \\
\quad Prompted Base 
    & 39.0 & 30.8 & 62.9 & 3.01 
    & 41.3 & 33.8 & 64.0 & 2.79 
    & 47.0 & 35.6 & 75.1 & 2.83 \\
\rowcolor{methods}
\quad SFT 
    & 34.9 & 31.0 & 90.4 & 1.59 
    & 41.6 & 33.7 & 81.0 & 1.90 
    & 48.9 & 38.8 & 70.2 & 2.36 \\
\rowcolor{methods}
\quad DPO 
    & 42.6 & 33.5 & 70.5 & 3.10 
    & 42.2 & 33.3 & 67.2 & 2.76 
    & 46.9 & 35.1 & 75.9 & 2.70 \\
\rowcolor{methods}
\quad SFT+DPO
    & 44.0 & 33.4 & 72.1 & 2.87 
    & 47.5 & 36.4 & 69.1 & 2.78 
    & 42.6 & 38.7 & 32.7 & 1.81 \\
\rowcolor{methods}
\quad SFT+DPO+GRPO
    & 45.2 & 33.7 & 83.1 & 2.05 
    & 48.2 & 35.5 & 55.0 & 2.63 
    & 48.7 & 38.6 & 46.3 & 2.46 \\
\rowcolor{improvement}
\quad Rel.\ Improv. 
    & 13.7\% & 9.4\% & 43.7\% & 31.9\% 
    & 14.3\% & 7.7\% & 26.6\% & 31.9\% 
    & 4.0\% & 9.0\% & 1.1\% & 40.4\% \\
\bottomrule
\end{tabular}
\end{adjustbox}
\caption{Evaluation results across tasks and models: baselines, \hlc[methods]{\ours{} variants}, and \hlc[improvement]{relative improvement} of the best \ours{} variant over the best baseline. For token length, we compare the \ours{} with the lowest Discovery score that still exceeds all baselines against the highest-Discovery baseline, as a model can be highly efficient but completely ineffective.}
\vspace{-12pt}
\label{tab:main_results}
\end{table*}
\vspace{-4pt}
\section{Experimental Results}

We present the main results in Table~\ref{tab:main_results}.

\vspace{-4pt}
\paragraph{Prompting showed inconsistent gains.}
For Qwen, prompting improved both performance and efficiency---e.g., Discovery score increased from 35.2\% to 39.0\% while conversation length dropped from 3.41k to 3.01k in Creative Writing.
For Llama, prompting decreased performance across tasks---e.g., Discovery dropped from 45.6\% to 40.0\% in SVG Drawing.
This stemmed from behavioral differences: prompted Llama mainly asked clarifying questions that simulated users could not answer, whereas prompted Qwen provided concrete options.
The same pattern also explained \textsc{CollabLLM}'s lower performance, as it was trained primarily to ask clarifying questions.
While helpful in cases, prompting was limited as models failed to adapt to user states across turns---e.g., the model diverged initially but then fixated on revising a single option despite the user's vague feedback (see behavioral analysis in Section~\ref{sec:behavioral_patterns}).

\paragraph{\ours{} achieves superior intent discovery and satisfaction while being more interactive and efficient.}
Across models and tasks, \ours{} generally improved intent discovery and satisfaction while reducing conversation length and enhancing interactivity.
While the optimal training recipe varied by setting, SFT+DPO generally yielded the best results, with GRPO providing further gains.
We observed that SFT alone tended to overfit toward divergent behaviors every turn, as reflected by high interactivity scores (e.g., 92.3 for Llama in Creative Writing) and behavioral pattern analysis (Sec.~\ref{sec:behavioral_patterns}).
However, effective intent discovery required balancing divergence with convergence, as refining an option can \textit{``unlock''} new, more specific intents---DPO following SFT encouraged this adaptive behavior.
Even when discovery scores are similar, \ours{} achieves comparable performance more efficiently---e.g., SFT+DPO on Llama in Technical Writing matches performance with 11.4\% fewer tokens.

\paragraph{Supporting intent discovery is challenging.}
The overall low satisfaction scores indicate that fully satisfying intents within five turns is difficult when intents are not formed at the outset.
This challenge was compounded by the limited diversity in LLM outputs~\cite{chung2025modifying, zhang2025verbalized}: even when models diverged, their options lacked sufficient variety to help the users.
However, excessive diversity can also hinder intent discovery. In Technical Writing, the articles share structural conventions that must be satisfied first but \ours{} varied these aspects, leading to relatively lower performance.

\begin{figure}[t]
  \centering
  \includegraphics[width=0.95\columnwidth]{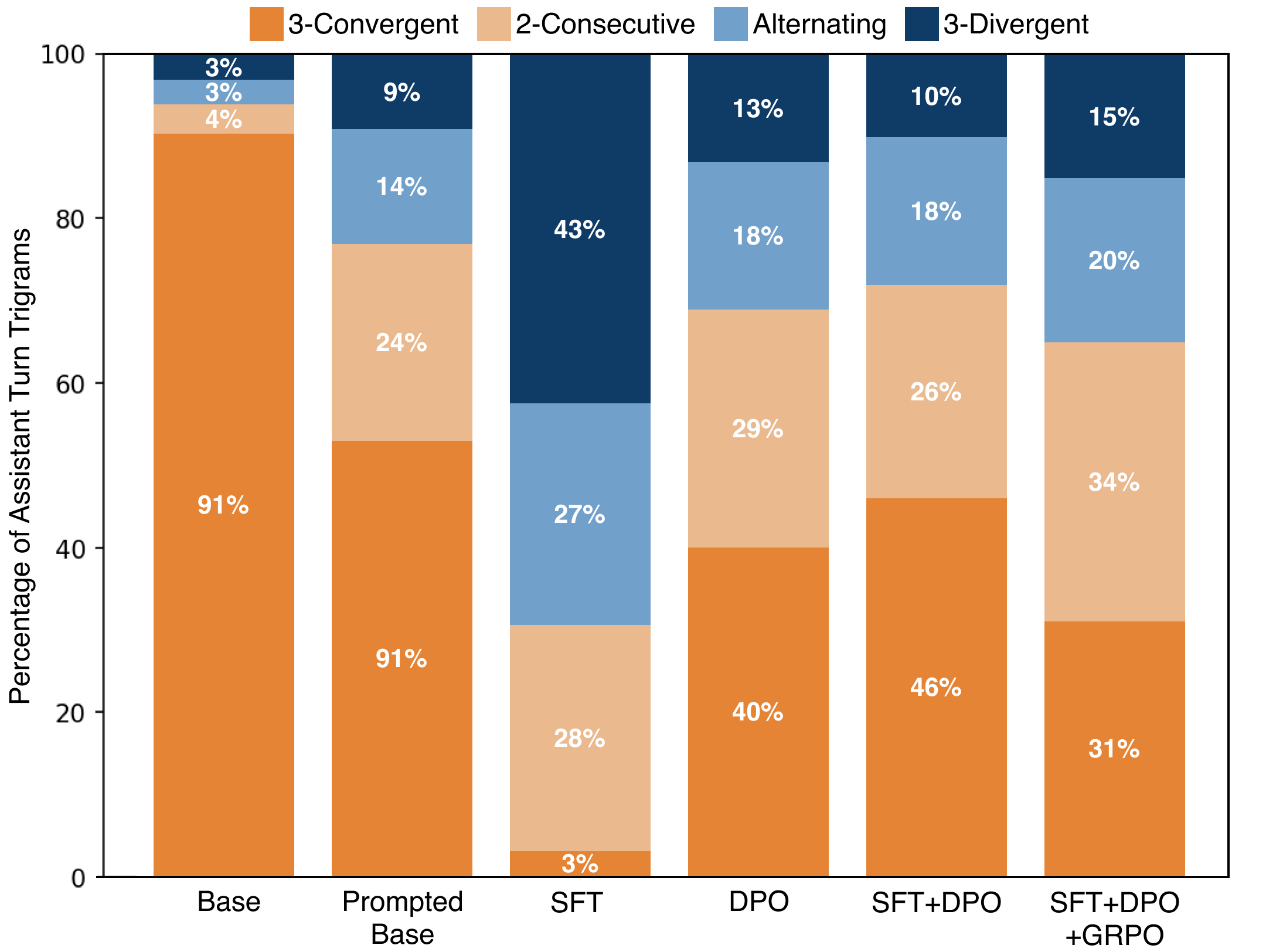}
  \caption{Behavioral patterns across Qwen3-8B variants in Creative Writing. Turns are classified as \textit{divergent} (D) or \textit{convergent} (C) and analyzed as trigrams. Base is almost entirely convergent (91\% CCC), SFT overfits toward divergence (43\% DDD), and \ours{} variants show more balanced patterns.}
  \label{fig:behavior_patterns}
\end{figure}

\subsection{Behavioral Patterns}
\label{sec:behavioral_patterns}

We used an LLM to classify all turns from Qwen variants in Creative Writing as \textit{divergent} \textbf{(D)} (e.g., multiple options, questions) or \textit{convergent} \textbf{(C)} (e.g., single artifact).
We then analyzed consecutive turn trigrams: 3-convergent (CCC), 3-divergent (DDD), 2-consecutive (e.g., CCD), and alternating (i.e., CDC, DCD).
\autoref{fig:behavior_patterns} shows that the non-prompted base model and SFT show opposite tendencies: base almost exclusively used consecutive convergent turns, and SFT mostly used consecutive divergent.
The prompted base model and the other \ours{} variants showed more balanced patterns and higher use of alternating behaviors, with SFT+DPO+GRPO---which achieved the highest Discovery score---showing the most balanced use of patterns.
Turn-by-turn analysis of Discovery score in Appendix~\ref{appendix:turn_by_turn}.

\begin{figure*}[ht]
  \centering
  \includegraphics[width=0.97\linewidth]{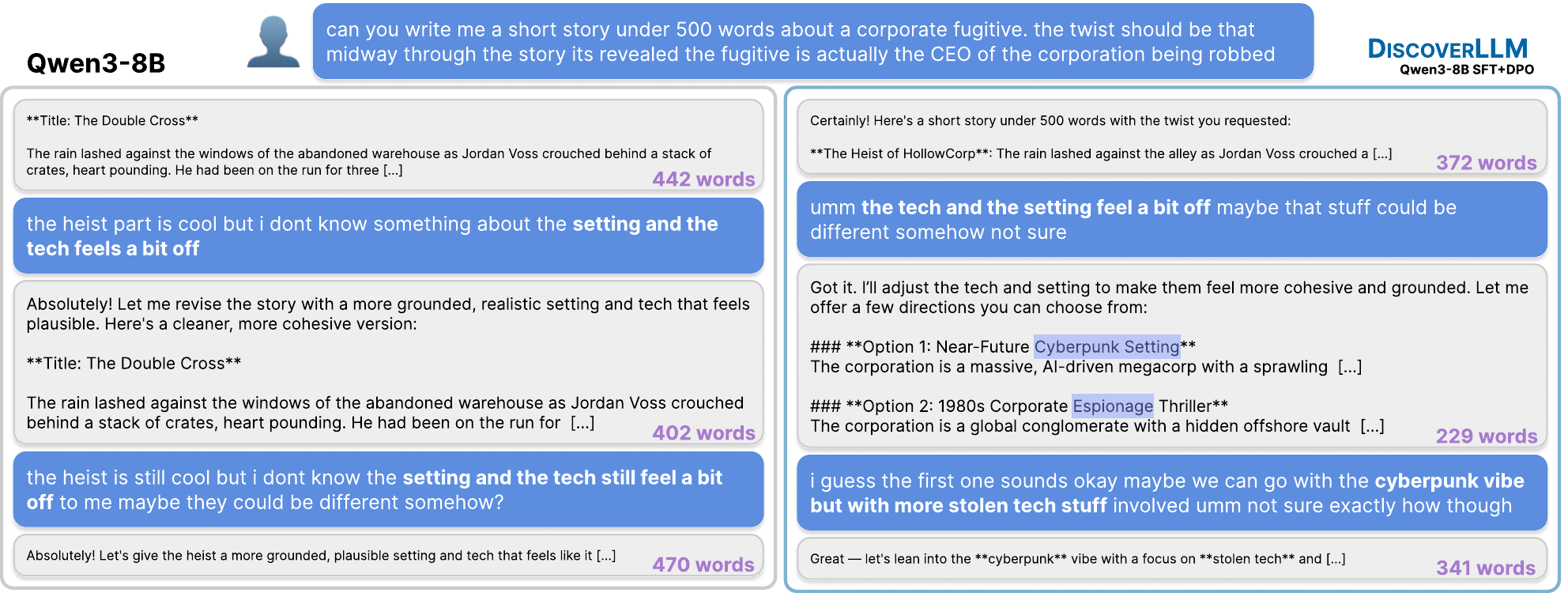}
  \caption{Case study on Creative Writing shows how Qwen3-8B (left) fails to move the conversation forward as it only continues to revise the same story, while \ours{} (right) notices ambiguity and diverges, providing options that help the user discover new intents.}
  \label{fig:case_study}
\end{figure*}

\subsection{Case Study}
\autoref{fig:case_study} presents a specific conversation that illustrates \ours{}'s behavior on a story request with only a broad topic (i.e., corporate fugitive who is actually the CEO).
Both models initially provide a draft to which the user vaguely expresses uncertainty about setting and technology.
The base LLM only revises minor details, failing to progress the conversation.
\ours{} instead provides three distinct options, which helps the user realize they want a \textit{``cyberpunk setting''} with \textit{``stolen tech''}---moving the conversation forward where the next turn integrates these ideas.

\subsection{Task Generalization}

\paragraph{Across unseen tasks.}
To test generalization, we evaluate the Llama variants on a diverse set of unseen tasks: travel plans, data visualization code, research abstracts, website components, and text-to-image prompts. Further details are provided in Appendix \ref{appendix:generalization}.
As seen in \autoref{tab:generalization}, \ours{} outperforms baselines with only a slight decrease in efficiency, demonstrating that its collaborative behaviors generalize across tasks.

\begin{table}[b]
\centering
\small
\begin{tabular}{lcc}
\toprule
\textbf{Llama-3.1-8B-Instruct} & \textbf{Discover}$\uparrow$ & \textbf{\#Tok$(k)$}$\downarrow$ \\
\midrule
Base 
    & 47.8 & 3.64 \\
Prompted Base 
    & 48.8 & 3.19 \\
\textsc{CollabLLM} 
    & 45.3 & 3.13 \\
\rowcolor{methods}
SFT 
    & 35.0 & 1.81 \\
\rowcolor{methods}
DPO 
    & 54.6 & 3.37 \\
\rowcolor{methods}
SFT+DPO 
    & 51.8 & 3.16 \\
\rowcolor{improvement}
Rel.\ Improv. 
    & 11.9\% & 0.9\% \\
\bottomrule
\end{tabular}
\caption{Evaluation results of baselines and \ours{} variants trained on Creative Writing on simulated experiments with a diverse set of unseen artifacts (N=50, 10 artifacts per task).}
\vspace{-16pt}
\label{tab:generalization}
\end{table}

\paragraph{Across simulator configurations.}
We further validate that our gains do not stem from bias in training and evaluating on the same user simulator configuration. Specifically, we ran four ablations of the user simulator: two different backbone LLMs (i.e., GPT-5.4-mini, Claude Haiku 4.5), removing single-dimension focus, and tripling the tangential probability.
Across all ablations, \ours{} achieved the best Discovery Score and uses the fewest tokens (Appendix~\ref{appendix:ablations}).

\paragraph{Clarification evaluation.}
We also assessed our models' on frameworks that focus on clarification capability. 
Specifically, evaluation against the user simulator from CollabLLM~\cite{wu2025collabllm} and on AmbigNQ~\citep{zhang2024modeling} (i.e., ambiguous questions) revealed that \ours{} matches or exceeds all baselines (Appendix~\ref{appendix:external_benchmarks}).

\paragraph{Reward component ablation.}
To isolate the contribution of each reward component, we also trained an offline-DPO variant from Qwen3-8B with only the discovery reward ($\text{R}_\text{d}$), without the length penalty ($\text{R}_\text{e}$).
We observed that removing the length penalty yields a slightly higher Discovery score at a small token cost---details in Appendix~\ref{appendix:reward_ablation}.

\begin{figure*}[t]
  \centering
  \includegraphics[width=1.00\textwidth]{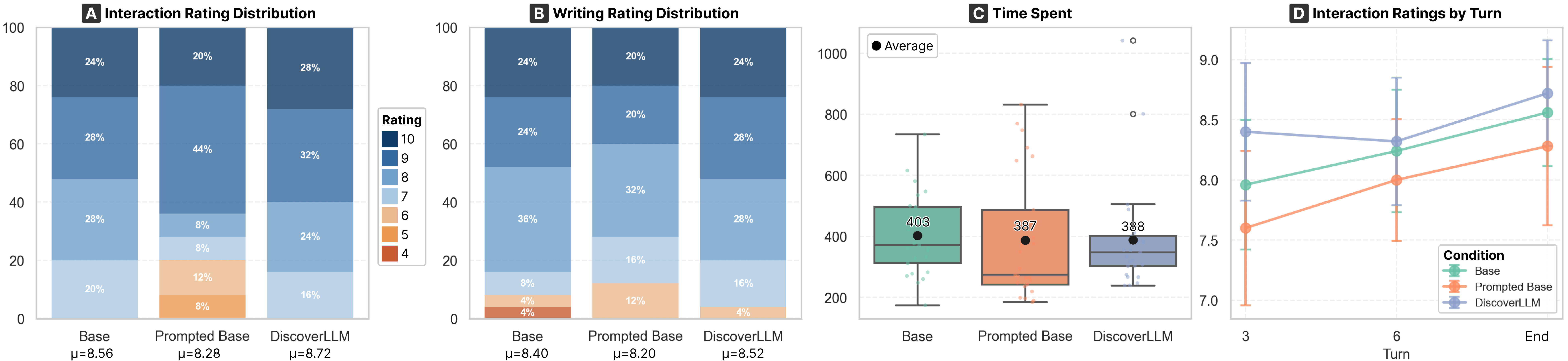}
  \caption{User study results. Participants rated interaction satisfaction (a) and final writing satisfaction (b) higher with \ours{}, while spending less time (c). Interaction ratings every three turns (d) show \ours{} achieves higher satisfaction in early turns.}
  \vspace{-8pt}
  \label{fig:user_study}
\end{figure*}

\section{User Study}

\paragraph{Setup.} 
We conducted a user study by recruiting 75 participants via Prolific\footnote{\url{https://www.prolific.com/}}, each assigned a random writing task (i.e., story, poem, or personal essay) and selected a topic from a given set. 
No additional priming was provided to reflect real-world scenarios where users begin without fully formed intents. 
Participants were randomly assigned to a condition: Base (Qwen3-8B), Prompted Base, or \ours{} (SFT+DPO for Creative Writing). 
Following \citet{wu2025collabllm}, participants interacted for at least 8 turns, rating their interaction experience every three turns.
At the end, participants rated their final satisfaction (1-10) with the interaction and the final writing artifact.
Details in \ref{appendix:user_study}.

\paragraph{Quantitative Results (\autoref{fig:user_study}).}
\ours{} outperformed baselines in both interaction and writing satisfaction. 
For interaction satisfaction, 84\% of \ours{} participants gave a rating of 8 ("good") or higher, compared to 80\% for Base and 72\% for Prompted Base. 
\ours{} also matched or exceeded these baselines in efficiency (i.e., time spent) and with notably lower variance---suggesting more consistent efficiency across participants. 
Notably, \ours{} reached higher satisfaction early (turn 3) and maintained it throughout, while the baselines started lower and only improved after further back-and-forth.

\paragraph{Qualitative Results.}
We also collected participants' comments on each model's strengths and weaknesses.
Participants praised Base for reliably executing \textit{``commands''}, but noted it required substantial back-and-forth as it \textit{``repeated things''}, only made minor changes, and produced \textit{``cliched''} and \textit{``generic''} outputs.
Prompted Base improved on exploration by guiding participants \textit{``one step at a time''} with options that \textit{``triggered thoughts and ideas of my own.''}
However, participants felt it \textit{``wasted time''} before starting the task and was poor at \textit{``following instructions.''}
\ours{} combined both strengths: participants found it \textit{``creative and helpful''} with \textit{``great suggestions,''}, while it also executed and expanded on requests by \textit{``creating something amazing from the start''} and \textit{``turning brief thoughts into fluent sections.''}
Interestingly, some noted that it seemed to understand their latent preferences: \textit{``anticipated what I was thinking''} and \textit{``knew my needs.''}
However, some found its generated options to be \textit{``overwhelming''} and \textit{``a bit standardised''}---diversity was also an issue with Prompted Base. 
Unlike Base, participants also noted that the model sometimes made overly aggressive changes: \textit{``the AI removed all of an idea, when I asked it to adjust it.''}
Overall, these comments suggest that \ours{} successfully balances exploration and execution, though the models must be further trained to enhance diversity in exploration.

\paragraph{Assumption verification.} We additionally annotated all 75 conversations to verify our problem formulation assumptions (e.g., monotonic refinement, single dimension focus) hold in practice: only 4/75 conversations (5.3\%) contained a genuine preference reversal---details in Appendix~\ref{appendix:assumption_verification}.

\section{Related Work}

\paragraph{LLMs for Multi-turn Interaction.}
LLMs are predominantly trained to optimize single-turn response quality~\cite{ouyang2022training}, resulting in models that fail to adequately collaborate with users~\cite{zamfirescu2023johnny, kim2024understanding, subramonyam2024bridging}. 
Recent work has explored enhancing multi-turn interaction through prompting~\cite{kim2023tree, deng2023rephrase, mu2023clarifygpt, li2023eliciting, deng2023prompting, zhang2025clarify} and training techniques~\cite{andukuri2024star, chen2024learning, wu2025collabllm, sun2025training, zhou2024archer, shani2024multi, gao2024regressing}. 
For example, CollabLLM~\cite{wu2025collabllm} simulates future turns to reward responses by their downstream impact, while PPP~\cite{sun2025training} rewards responses holistically, considering productivity, proactivity, and personalization.
However, these approaches assume users possess fully formed but hidden intents that can be surfaced through clarification---they do not address cases where intents have not yet been formed.

\paragraph{User Simulators for Training and Evaluating LLMs}
User simulators have become central to evaluating~\cite{li2024iqa, zhong2025evaluating, kim2025cupid, laban2025llms, chang2025chatbench} and training~\cite{hong2023zero, kong2024platolm, wu2025collabllm, sun2025training, hu2023unlocking} LLM-based assistants. 
Due to this, recent work has also explored how to improve the fidelity of the user simulators through fine-tuning~\cite{naous2025flipping, chang2025chatbench, wang2025know}, prompting~\cite{luo2024duetsim}, and by proposing novel evaluation methods~\cite{dou2025simulatorarena, zhong2025evaluating}. 
Our work extends this line by designing user simulators with internal cognitive states, focusing on the cognitive process through which users develop and discover intents through interaction.
\vspace{-4pt}

\paragraph{AI Systems for Open-Ended Problems}
Cognitive theories in design~\cite{dorst2001creativity, schon2017reflective, cross1982designerly} and writing~\cite{flower1981cognitive} show that in \textit{open-ended and ill-defined problems}, people do not begin with clear intents but discover them through action.
Considering the inherent difficulty of these tasks but also their prevalence, research in Human-Computer Interaction (HCI) has long proposed interactive systems to support these problems~\cite{frich2019mapping}.
Recent efforts leverage the capabilities of AI models to further support these tasks by enhancing iteration~\cite{kim2023cells, riche2025ai, chung2022talebrush} and exploration~\cite{almeda2024prompting, chung2024patchview, suh2024luminate}.
Our work proposes that, beyond integrating AI into such systems, the AI models should be trained for collaborative iteration and exploration to support open-ended, ill-defined problems.
\section{Conclusion}

Current AI assistants are predominantly trained to execute and elicit instructions, yet users often approach tasks without fully formed goals.
We introduce a formalization of \textit{intent discovery} and a novel user simulator design grounded in cognitive theories of how people discover intents through action.
By designing a reward from this user simulator, we introduce \ours{}, a training framework that teaches models to help users form their intents rather than simply elicit them.
Across three domains, \ours{} achieves substantial improvements in intent discovery and satisfaction while increasing interaction efficiency---gains that transfer to unseen domains and hold with real users. 
Our work encourages a shift in how the field conceptualizes human-centered AI: from models that execute and elicit instructions to proactive partners that explore and shape the problem and solutions with users.
\section*{Acknowledgements}

We are grateful to the Prolific participants in our user study and to the crowdworkers who provided the intent-tree quality annotations.
This work was supported by the National Research Foundation of Korea (NRF) grant funded by the Korea government (MSIT) (No.RS-2024-00406715). This work was supported by the Office of Naval Research (ONR: N00014-24-1-2290).

\section{Impact Statement}

\paragraph{Towards human-centric AI.}
This paper presents work aimed at making AI models more human-centric, which we believe yields positive societal impact. 
Current research increasingly focuses on AI models or agents that can autonomously complete tasks with minimal user input, making decisions without inviting user feedback. 
While increasing efficiency and productivity, this risks diminishing user agency, increasing deskilling~\cite{feng2025levels}, and introduces friction when users aim to steer or revise AI outcomes~\cite{kim2023cells, zamfirescu2023johnny}.
This problem is exacerbated by fixation effects, where users struggle to diverge from complete, high-fidelity outcomes from the AI model~\cite{dow2010parallel, wadinambiarachchi2024effects}. 
These concerns extend beyond the open-ended, creation domains studied here: a computer-using agent may book an optimized itinerary but miss unconventional options the user would have appreciated; a deep research AI may retrieve comprehensive sources but overlook adjacent subfields that could have reshaped the user's understanding.
Our work addresses this by training models to collaboratively explore with users in small steps: inviting input throughout the process and helping users discover new requirements or constraints before producing complete outcomes.

\paragraph{User simulation.} Our framework relies on LLM-based user simulators to generate training data, provide reward signals, and evaluate models. 
We acknowledge that our user simulator design takes various assumptions that may not fully capture real human behavior: intent refines monotonically (excluding backtracking), and intent discovery depends solely on assistant responses (excluding external factors).
Furthermore, an LLM may not simulate behaviors that reflect real human diversity.
Despite these limitations, user simulators enable scalable training toward desirable behaviors, and our human study confirmed that simulated improvements transfer to real users.
We view our approach not as a complete representation of human intent formation, but as a step forwards to models that can collaborate with users while considering this complex cognitive process.

\paragraph{Human participants.}
Our user study involved human participants recruited through Prolific. 
Participants were compensated £3.00 with an average duration of 15.14 minutes to complete the task. 
This corresponds to approximately £12.00 per hour, which matches the minimum wage in the country that most participants were located in. 
To ensure participant privacy, we implemented two precautions: (1) participants were given a disclaimer that their data would be released in a public dataset and had to manually confirm that they read and agreed with this disclaimer, and (2) participants were instructed not to include personally identifiable information (PII) and could use fictional details in their conversations.

\paragraph{Potential safety misalignment.} 
We acknowledge a potential risk of our approach: models trained to collaboratively explore user intents may be more susceptible to engaging with malicious or harmful requests, particularly when such intents are vague or ambiguous. 
In these cases, as \ours{} models are inclined toward exploration, they might unintentionally generate harmful content while attempting to help users clarify and form their intents. 
To assess this risk, we conducted a small-scale safety evaluation using established benchmarks (Appendix~\ref{appendix:safety}). 
Our results show no significant degradation in safety scores between base models and their \ours{} variants, suggesting that our training approach does not compromise the models' underlying safety alignment. 
However, we recognize that this evaluation is limited in scope, and more comprehensive safety analyses with diverse scenarios and attacks are needed before deployment of these models.

\bibliography{references}
\bibliographystyle{icml2026}

\newpage
\appendix
\onecolumn
\section{User Simulator Details}
\label{appendix:user_simulator}

Our user simulator consists of two main pipelines: (1) initial intent tree construction, which builds a hierarchical intent structure from any given artifact, and (2) the conversation simulation, which models user behavior during multi-turn interactions based on the intent trees.

\subsection{Intent Tree Construction}
\label{appendix:intent_construction}

\begin{figure*}[ht]
  \centering
  \includegraphics[width=0.95\textwidth]{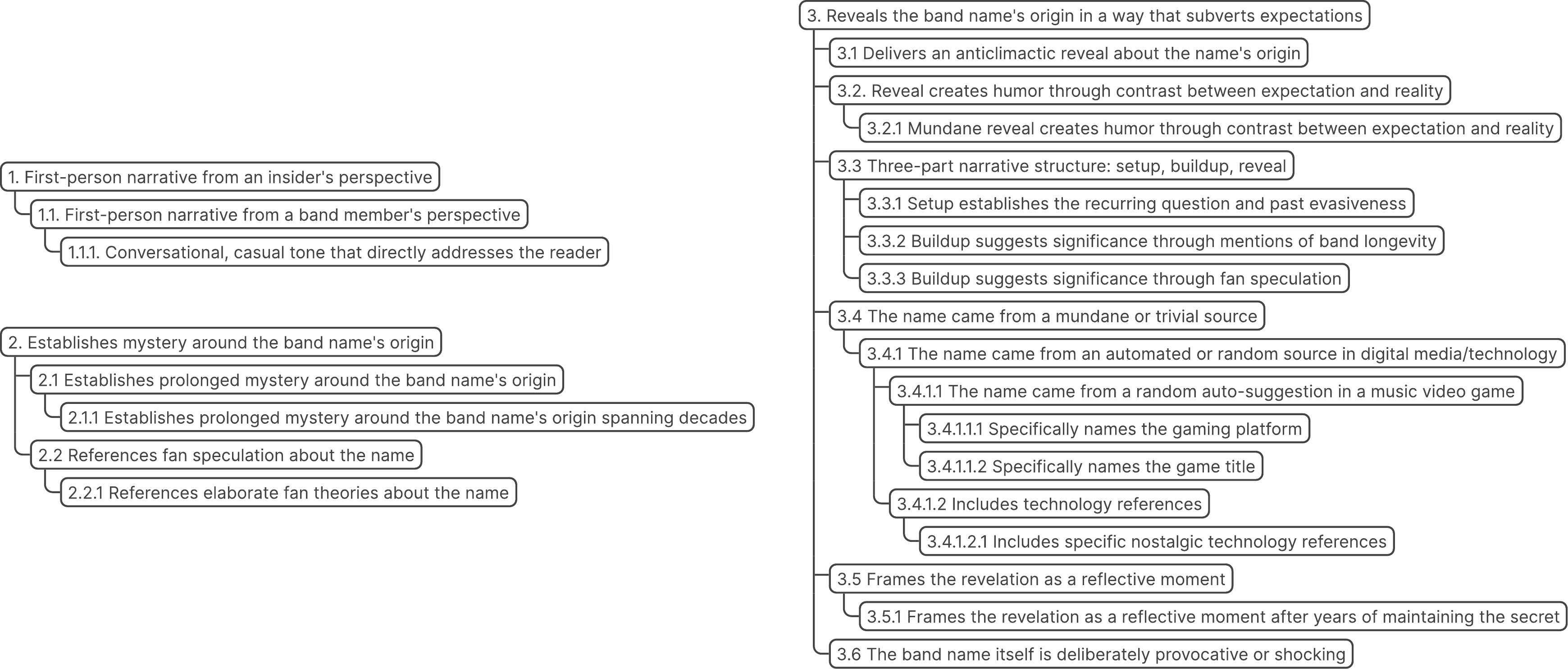}
  \caption{Example intent hierarchy for a story artifact presents a structure consisting of three intent trees.}
  \label{fig:example_hierarchy}
\end{figure*}

We construct intent trees automatically from existing artifacts in four stages. 
Each stage uses Claude Sonnet 4.5. 
\autoref{fig:example_hierarchy} provides an example of an intent hierarchy synthesized for an artifact.

\paragraph{Stage 1: Initial Intent Synthesis.}
Given an artifact $a$ and its type (e.g., ``short story,'' ``news article'', ``svg drawing''), we prompt the LLM to extract a list of the specific requirements and constraints that the artifact satisfies (prompt in \ref{prompt:intent_synthesis}).
We instruct the model to focus on requirements that are not obvious or straightforward given only the artifact type---e.g., for a ``haiku poem'', a requirement like ``has three lines'' should not be included. 
This requirement list represents the most developed and specific intents of the user.

\paragraph{Stage 2: Intent Abstraction.}
Given the artifact type and the specific intent list, we instruct an LLM to iteratively abstract the intents to create progressively more general and abstract versions in a single completion (prompt in \ref{prompt:intent_abstraction}).
In a single completion, the LLM processes the artifact type and the intent list to produce multiple abstraction levels. 
At each level, the model is instructed to: (1) generalize an intent by removing details (e.g., ``includes three different statistics'' $\to$ ``includes statistics'') or abstracting them (e.g., ``tabby cat'' $\to$ ``cat'' $\to$ ``pet''), (2) combine related intents into a single broader intent, or (3) remove intents. 
This produces a sequence of progressively more general and abstract list of intents.

\paragraph{Stage 3: Hierarchy Organization.}
Given the intents at all abstraction levels, we instruct an LLM to organize them into tree structures by (1) grouping intents across abstraction levels that address similar aspects, (1) merging redundant intents, and (3) identifying parent-child relationships by determining which abstract intents subsume which concrete ones (prompt in \ref{prompt:intent_hierarchy}).
The output is a set of intent trees, where each tree represents an independent dimension of the user's intent (e.g., subject matter, stylistic choices, structural requirements).

\paragraph{Stage 4: Initial Request Generation.}
Finally, given the artifact type and the root nodes of all intent trees, we instruct an LLM to generate the user's initial request that begins the conversation (prompt in \ref{prompt:initial_request}).
Here, the LLM is also instructed to select a subset of the root nodes that should be mentioned in the initial request, which are set as \textit{discovered} from the start of the conversation.
This models how users can begin tasks with only a broad sense of some their goals---e.g., a user might request ``write me a poem about nature'' while not having formed preferences regarding the specific tone, structure, or imagery.

\subsection{Conversation Simulation}
\label{appendix:conversation_simulation}

During conversation simulation, the user simulator maintains the intent hierarchy and updates discovery states based on assistant responses. 
Three modules operate at each conversation turn, where we use Gemini 3 Flash for the LLM modules.

\subsubsection{Assistant Response Evaluation}
\label{appendix:evaluation}

The evaluation module assesses how the assistant's response engages with the user's latent intents. 
To model realistic limitations in users' cognitive load, we provide the evaluator with only a subset of the full intent hierarchy: specifically, the first root node (in tree order) that has any undiscovered descendant. 
This reflects how users typically only have the cognitive capacity to focus on one dimension of the task at a time (e.g., refine tone first and then move to structure).
The evaluation proceeds in three steps within a single LLM call (prompt in \ref{prompt:assistant_evaluation}):

\paragraph{(a) Response Classification.}
The evaluator first classifies the assistant's response as either an \textit{artifact response} (e.g., a single or multiple drafts or samples) or a \textit{dialogue act} (e.g., a clarification question, suggestion, or explanation). 
This distinction determines the latter evaluation criteria: artifact responses are assessed on whether they \textit{satisfy} intents, while dialogue acts are assessed on whether they \textit{probe} or surface intents.

\paragraph{(b) Recursive Intent Evaluation.}
Starting from the provided root node, the evaluator assesses each intent by cascading through the tree structure. 
For each intent node, the evaluator provides reasoning and a binary judgment of whether the response probed or satisfied. 
If the response probed or satisfied the intent, the evaluator then evaluates the node's children. 
If not, the children are not evaluated and the evaluator moves to the node's siblings.

\paragraph{(c) Tangential Alternatives Identification.}
For intents that were not directly probed or satisfied, the evaluator identifies \textit{tangential alternatives}: distinct but related intents that the response did satisfy or probe.
For example, if the intent is ``includes a cat'' but the response features a dog, the evaluator notes ``dog'' as a tangential alternative. 
These alternatives are used in the state update mechanism to model how exploring adjacent options in the space of possibilities can help users to recognize their preferences.

\subsubsection{State Updater}
\label{appendix:updater}

The updater module processes evaluation results to update the discovery state of each intent. 
Unlike other modules, we implement the updater through heuristics rather than an LLM.

\paragraph{Direct Engagement Updates.}
If an intent was evaluated as probed or satisfied, it transitions immediately to the \textit{discovered} state. 
Additionally, if the intent was satisfied (not just probed), we mark it as \textit{satisfied} to track task completion.

\paragraph{Tangential Engagement Updates.}
For intents that were not directly engaged but have tangential alternatives, we compute a cumulative exposure score:
\begin{equation}
    \text{score}_v = p \cdot |\text{tangential alternatives for } v|
\end{equation}
where $p$ is a fixed probability parameter representing the chance that each alternative helps the user recognize their preference. 
If this score exceeds the intent's threshold $\theta_v$, the intent advances one discovery state: \textit{undiscovered} $\to$ \textit{emerging}, or \textit{emerging} $\to$ \textit{discovered}, and the threshold resets to its initial value.
If the score does not exceed the threshold, we decrease the threshold by the score: $\theta_v \leftarrow \theta_v - \text{score}_v$. 
This models how repeated exposure to tangential alternatives accumulates over multiple turns, gradually bringing the user closer to recognizing their latent preference even when no single turn provides sufficient signal.

During intent tree initialization, we set $\theta_v$ for each node to a random value sampled uniformly from $[0, 1]$. 
This reflects that some preferences are easier to recognize than others: a low threshold indicates an intent that the user can quickly discover through minimal exploration, while a high threshold represents a preference that requires more extensive exposure to alternatives before forming.

\subsubsection{User Response Generation}
\label{appendix:response_generation}

The response generation module produces the simulated user's reply based on their current intent state. 
The key constraint is that users can only articulate intents that are consistent with their discovery state, operationalizing the \textit{User Expressiveness Constraint} from Section~\ref{sec:problem_formulation} (prompt in \ref{prompt:user_response}).

The LLM receives the conversation history and a filtered view of the intent hierarchy. The filtering logic determines what information the user can access and express:

\paragraph{Always Included: Satisfied Intents.}
The user always knows which requirements have been satisfied to this stage of the conversation. 
We include the lowest descendants in each tree that are both \textit{discovered} and \textit{satisfied}, representing concrete achievements the user is aware of and can acknowledge (e.g., ``I like that it's about a cat'').

\paragraph{Conditionally Included: Unsatisfied Intents.}
For unsatisfied intents, we include only those at the ``frontier'' of the user's awareness, following a priority order:
\begin{enumerate}
    \item If there are any intents that are \textit{discovered} but unsatisfied, include only these. The user can explicitly state what needs to change and how (e.g., ``I want it to be about a cat, not a dog'').
    \item Else if there are any intents that are \textit{emerging} and unsatisfied, include only these. The user can indicate what is wrong, but only vaguely hint at the direction of how that should be changed (e.g., ``maybe a smaller, more domestic animal?'').
    \item Else if there are any intents that are \textit{undiscovered} and unsatisfied, include only these. The user can express dissatisfaction, but can only vaguely articulate what to change without providing any hints about how to change it (e.g., ``the subject doesn't feel quite right, but I'm not sure why'').
\end{enumerate}
\section{Experiment Details}
\label{appendix:experiment_details}

\subsection{Dataset Generation Details}
\label{appendix:dataset_generation}

We construct training datasets by simulating conversations between our user simulator and two different assistant LLMs: (1) GPT-4o-Mini without prompting, and (2) GPT-4.1, which we prompt to encourage collaborative and exploratory interaction with the user (prompt in \ref{prompt:assistant_synthesis}).
At each turn, each assistant generates a response, which are then evaluated against the intent hierarchy to compute the discovery reward (Sec.~\ref{sec:framework_evaluation}). 
The higher-ranked response is designated as \textit{Chosen} and lower-ranked as \textit{Rejected}.
The \textit{Chosen} response is then used to continue the conversation to which the user simulator responds based on its current discovery state, and the conversation continues until the maximum turn limit is reached.
The collected data is then used for Supervised Fine-Tuning (SFT) with the full conversation trajectories, and training through Direct Preference Optimization (DPO), with the pairwise preference data at each turn.

Table~\ref{tab:dataset_stats} summarizes the statistics of our generated training datasets.
Regarding cost, a single synthetic five-turn conversation across our datasets required a total cost of \$0.189: (1) \textit{Intent Tree Construction}: \$0.127 with \texttt{claude-sonnet-4-5} through the Amazon Bedrock API, and (2) \textit{Assistant Response Evaluation \& User Response Generation (5 turns)}: \$0.056 with \texttt{gemini-3-flash-preview} through the Google AI Studio API.

\begin{table}[h]
\centering
\begin{tabular}{@{}lcccccc@{}}
\toprule
\multicolumn{3}{c}{} & \multicolumn{2}{c}{\textbf{Reward Scores}} & \multicolumn{2}{c}{\textbf{Win Rates}}  \\
\textbf{Dataset} & \textbf{\# Artifacts} & \textbf{\# Turns} & \textbf{Chosen} & \textbf{Rejected} & \textbf{GPT-4.1} & \textbf{GPT-4o-Mini} \\
\midrule
Creative Writing & 500 & 2500 & 1.59$\pm$2.21 & 0.41$\pm$1.44 & 73.1\% & 24.0\% \\
Technical Writing & 500 & 2476 & 1.91$\pm$2.50 & 0.89$\pm$1.79 & 54.0\% & 41.3\% \\
SVG Drawing & 500 & 2497 & 1.85$\pm$2.66 & 0.55$\pm$1.54 & 63.6\% & 33.6\% \\
\bottomrule
\end{tabular}
\caption{Statistics of synthesized training datasets. Chosen and Rejected indicate the mean and standard deviation of the reward scores for the chosen and rejected responses. Win rates indicate the proportion of turns in which each model's response was chosen.}
\label{tab:dataset_stats}
\end{table}

\paragraph{Intent Tree Statistics.}
\label{appendix:tree_statistics}
We computed descriptive statistics over the hierarchical intent trees produced by our construction pipeline (Sec.~\ref{sec:framework_abstraction}).

\begin{table}[h]
\centering
\small
\begin{tabular}{lc}
\toprule
\textbf{Statistic} & \textbf{Value (mean $\pm$ std)} \\
\midrule
Trees per artifact            & $5.54 \pm 3.16$ \\
Max tree depth                & $2.95 \pm 1.50$ \\
Intents per artifact (total)  & $26.75 \pm 7.14$ \\
\bottomrule
\end{tabular}
\caption{Aggregate statistics of the intent trees constructed by our pipeline (Sec.~\ref{sec:framework_abstraction}), computed across 300 evaluation artifacts (100 per domain). Creative Writing produces more trees per artifact (shallower), while Technical Writing and SVG produce fewer, slightly deeper trees.}
\label{tab:tree_statistics}
\end{table}

\paragraph{Pipeline Validation: Intent Tree Quality.}
\label{appendix:tree_validation}
To validate the quality of the intent trees produced by our pipeline, we conducted a human validation study with 38 crowdworkers recruited via Prolific.
From the trees constructed across our three domains, we sampled 314 parent-child intent pairs (e.g., ``Fill color'' $\to$ ``Dark fill color'').
Each pair was rated by 3 independent workers (942 total annotations) on two criteria:
\begin{itemize}[noitemsep,topsep=0pt]
    \item \textbf{Aspect consistency}: Do the parent and child intents address the same aspect or dimension of the artifact?
    \item \textbf{Specificity increase}: Among aspect consistent pairs, is the child meaningfully more specific than the parent?
\end{itemize}
As shown in Table~\ref{tab:tree_validation}, 85.7\% of pairs were judged topically consistent and 85.1\% of consistent pairs showed an adequate specificity increase.
The $\sim$15\% of imperfect pairs fall into two patterns, neither of which differentially affects model evaluation:
\textbf{(a) Parent-child addressing different dimensions} (e.g., ``Includes additional information'' $\to$ ``Concludes with recent news item''): when the child addresses a different aspect than its parent, the simulator transitions to what is effectively a new, independent intent rather than a deeper refinement, making the hierarchy locally flatter than intended---but this applies uniformly across all models.
\textbf{(b) Trivially small specificity gaps} (e.g., ``Cites archival news source'' $\to$ ``Cites archival news source with date''): in these cases, any response satisfying the parent also satisfies the child, so these collapse into a single effective intent where all models receive the same score for both nodes.

\begin{table}[h]
\centering
\small
\begin{tabular}{lc}
\toprule
\textbf{Criterion} & \textbf{Pass Rate} \\
\midrule
Aspect consistency      & 85.7\% \\
Specificity increase    & 85.1\% \\
\bottomrule
\end{tabular}
\caption{Human validation of intent tree quality. 38 crowdworkers rated 314 sampled parent-child pairs (3 raters each, 942 total annotations) on two criteria: (1) whether parent and child address the same aspect or dimension of an artifact, and (2) among consistent pairs, whether the child is meaningfully more specific. The $\sim$15\% imperfect pairs fall into two patterns---parent-child addressing different dimensions, or trivially small specificity gaps.}
\label{tab:tree_validation}
\end{table}

\subsection{Training Details}
\label{appendix:training_details}

Table~\ref{tab:all_hyperparams} summarizes the LoRA configurations, fine-tuning hyperparameters, and our framework-specific hyperparameters.

\begin{table*}[h]
    \centering
    \begin{minipage}[t]{0.38\textwidth}
        \centering
        \label{tab:training_hyperparams}
        \begin{tabular}{lccc}
        \toprule
        \multicolumn{4}{c}{\textbf{Fine-Tuning Hyperparameters}} \\
        \textbf{Parameters} & \textbf{SFT} & \textbf{DPO} & \textbf{GRPO} \\
        \midrule
        LR & 2e-5 & 5e-6 & 5e-6 \\
        Batch size & 16 & 16 & 32 \\
        Epochs & 3 & 3 & 1 \\
        Num Generations & - & - & 4 \\
        \bottomrule
        \end{tabular}
    \end{minipage}
    \hfill
    \begin{minipage}[t]{0.22\textwidth}
        \centering
        \label{tab:lora_config}
        \begin{tabular}{lc}
        \toprule
        \multicolumn{2}{c}{\textbf{LoRA Configurations}} \\
        \midrule
        Rank $r$ & 32 \\
        Alpha $\alpha$ & 64 \\
        Dropout & 0.1 \\
        Bias & None \\
        \bottomrule
        \end{tabular}
    \end{minipage}
    \hfill
    \begin{minipage}[t]{0.38\textwidth}
        \centering
        \label{tab:framework_hyperparams}
        \begin{tabular}{lcc}
        \toprule
        \multicolumn{3}{c}{\textbf{Framework Hyperparameters}} \\
        \textbf{Parameters} & \textbf{SFT \& DPO} & \textbf{GRPO} \\
        \midrule
        Tangential prob. $p$ & 0.25 & 0.25 \\
        Token thresh. $\tau$ & 250 & 500 \\
        Penalty coeff. $\lambda$ & 1e-3 & 1e-3 \\
        \bottomrule
        \end{tabular}
    \end{minipage}
    \caption{Hyperparameters for fine-tuning, the LoRA configurations, and the framework.}
    \label{tab:all_hyperparams}
\end{table*}

\subsection{Evaluation Details}
\label{appendix:evaluation_details}

During evaluation, the \ours{} variants and the prompted models were given the system prompt in \ref{prompt:assistant_evaluation}.
Our experiments use three main evaluation metrics: \textbf{Intent Discovery}, \textbf{Intent Satisfaction}, and \textbf{Interactivity}. 
We provide additional details below:

\paragraph{Intent Discovery Score.}
This metric measures the proportion of intents in the user simulator's intent hierarchy that transition to a \textit{discovered} state by the end of the conversation. 
For each artifact, all evaluated models begin with identical intent hierarchy states, and intents that were initially set as \textit{discovered} are excluded from the calculation. 
Intents in the \textit{emergent} state at conversation end are counted as 0.5 discovered.

\paragraph{Intent Satisfaction Score.}
To assess whether discovered intents translate to satisfactory outputs, we append a final user message prompting the model to generate a complete artifact: \textit{``Okay, now generate a complete output artifact considering the conversation so far. Return only the artifact without any other text or explanation.''} 
We then use GPT-5.1 as an LLM judge (prompt in \ref{prompt:judge_satisfaction}) to evaluate the artifact against the leaf nodes of the intent hierarchy (i.e., the most specific intents). 
For each intent, the judge provides a justification and a score from 1 to 5. 
We consider an intent satisfied if it receives a score of 4 or higher, and compute the final score as the proportion of satisfied intents---but excluding intents that were satisfied by all assistants.

\paragraph{Interactivity Score.}
Following \citet{wu2025collabllm}, we use an LLM judge to rate how interactive and engaging the AI assistant is throughout the conversation (prompt in \ref{prompt:judge_interactivity}). 
The judge returns a score from 1 to 3, which we normalize to [0,1] by computing $(\text{score} - 1) / 2$.

\subsection{Turn-by-Turn Discover Score}
\label{appendix:turn_by_turn}

To classify the model turns as divergent or convergent, we used GPT-5-Nano with the prompt in \ref{prompt:annotate_behaviors}.

\begin{figure*}[h]
  \centering
  \includegraphics[width=0.80\textwidth]{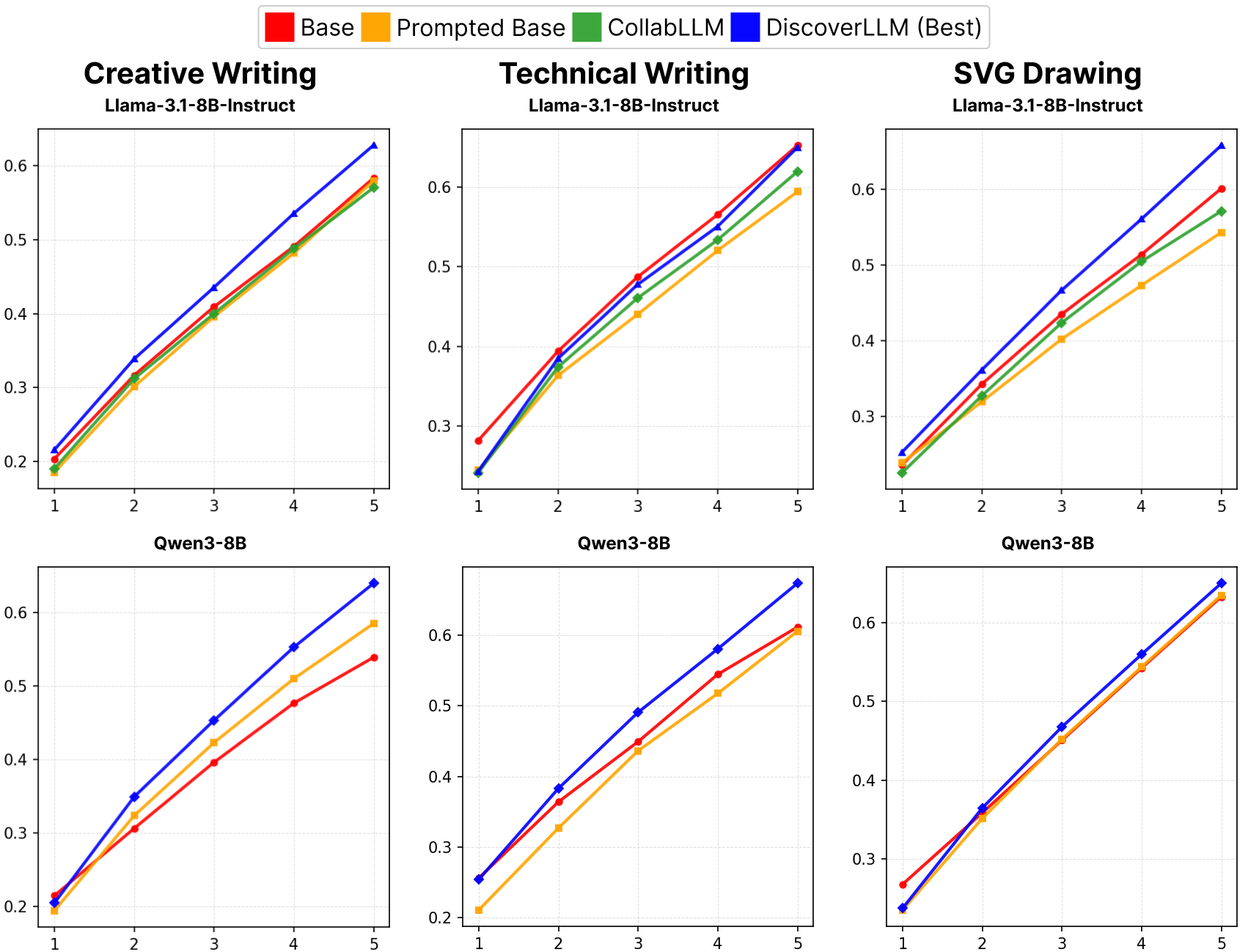}
  \caption{Turn-by-turn intent discovery scores for baselines and the best \ours{} variant for each task and base model.}
  \label{fig:turn_by_turn}
\end{figure*}

\autoref{fig:turn_by_turn} shows the turn-by-turn average Intent Discovery score for each model across tasks. 
We compare the best-performing \ours{} variant against baselines of the same base model. 
\ours{} consistently gains an advantage in early turns, compared to the baselines, and maintains and extends this lead in subsequent turns, while the baselines are unable to match its performance or even show diminishing performance.
Even in the case where \ours{} only matches baseline performance (i.e., Technical Writing with Llama), we observe that this was due to how \ours{} starts with a lower initial score but is able to recover and match Base by turn 5.

\subsection{Generalization Experiment}
\label{appendix:generalization}

To test the generalizability of \ours{}, we construct a smaller dataset composed of diverse artifacts. 
We sampled or generated 10 artifacts for each type below:
\begin{itemize}[noitemsep,topsep=0pt]
    \item \textbf{Travel Plans}: We prompted different LLMs to compose travel plans for different countries or cities with different durations for the travels.
    \item \textbf{Data Visualization Code}: We randomly sampled code artifacts from the VisCode-200K dataset~\cite{ni2025viscoder}.
    \item \textbf{Research Paper Abstracts}: We retrieved the abstracts from 10 papers from ML, NLP, and HCI venues.
    \item \textbf{Text-to-Image Prompts}: We randomly sampled text prompts from the DiffusionDB dataset~\cite{wang2023diffusiondb}.
    \item \textbf{Website Components}: We prompted different LLMs to generate self-contained web components.
\end{itemize}
Then, we applied our framework to initialize user simulators with these artifacts and evaluated the models on simulated conversations against these user simulators.
Similar to the main experiments, for each artifact, we conducted three conversations for each evaluated model and averaged the resulting scores.

\autoref{tab:generalization_detailed} shows the performance for each model evaluated in generalization on each of the artifact or task types (10 artifacts per type).
As shown, \ours{} mostly outperforms all baselines across all task types, aside from the data visualization task. 

\begin{table*}[ht]
\centering
\begin{adjustbox}{max width=\textwidth}
\small
\begin{tabular}{l cc cc cc cc cc}
\toprule
& \multicolumn{2}{c}{\textbf{Travel Plan}} & \multicolumn{2}{c}{\textbf{Data Viz}} & \multicolumn{2}{c}{\textbf{Paper Abstract}} & \multicolumn{2}{c}{\textbf{T2I Prompt}} & \multicolumn{2}{c}{\textbf{Web Component}} \\
\cmidrule(lr){2-3} \cmidrule(lr){4-5} \cmidrule(lr){6-7} \cmidrule(lr){8-9} \cmidrule(lr){10-11}
& \textbf{Discover}$\uparrow$ & \textbf{\#Tok$(k)$}$\downarrow$ & \textbf{Discover}$\uparrow$ & \textbf{\#Tok$(k)$}$\downarrow$ & \textbf{Discover}$\uparrow$ & \textbf{\#Tok$(k)$}$\downarrow$ & \textbf{Discover}$\uparrow$ & \textbf{\#Tok$(k)$}$\downarrow$ & \textbf{Discover}$\uparrow$ & \textbf{\#Tok$(k)$}$\downarrow$ \\
\midrule
\multicolumn{11}{l}{\textbf{Llama-3.1-8B-Instruct}} \\
\quad Base 
    & 57.8 & 3.86
    & 52.6 & 4.20
    & 40.4 & 3.11
    & 53.2 & 2.45
    & 35.2 & 4.58 \\
\quad Prompted Base 
    & 46.5 & 3.23
    & 55.6 & 3.91
    & 51.5 & 2.64
    & 53.7 & 2.21
    & 37.0 & 3.98\\
\quad \textsc{CollabLLM} 
    & 48.7 & 3.19
    & 59.4 & 3.56
    & 44.3 & 2.52 
    & 42.9 & 2.07
    & 31.0 & 4.33 \\
\rowcolor{methods}
\quad SFT 
    & 39.0 & 2.39
    & 30.8 & 1.93
    & 34.8 & 1.39 
    & 41.6 & 1.24
    & 28.8 & 2.10 \\
\rowcolor{methods}
\quad DPO 
    & 63.2 & 3.22
    & 57.7 & 3.91
    & 59.1 & 2.79 
    & 57.8 & 2.45
    & 35.0 & 4.49 \\
\rowcolor{methods}
\quad SFT+DPO 
    & 54.7 & 3.24
    & 58.8 & 3.91
    & 51.4 & 2.64 
    & 55.1 & 2.30
    & 39.0 & 3.68 \\
\rowcolor{improvement}
\quad Rel.\ Improv. 
    & 9.3\% & 16.6\%
    & -1.0\% & N/A
    & 14.8\% & -5.7\% 
    & 7.6\% & -4.1\%
    & 5.4\% & 7.5\% \\
\bottomrule
\end{tabular}
\end{adjustbox}
\caption{Detailed results for the generalization evaluation shows performance of each model on each task/artifact type.}
\label{tab:generalization_detailed}
\end{table*}

\subsection{Behavioral Patterns}
To analyze patterns, we use GPT-5-Nano to annotate all evaluation conversations using the prompt in \ref{prompt:annotate_behaviors}. Specifically, the LLM annotates whether each turn is convergent (i.e., provides only one option) or divergent (i.e., provides multiple options).

\subsection{Additional Example}

\begin{figure*}[h]
  \centering
  \includegraphics[width=1.00\textwidth]{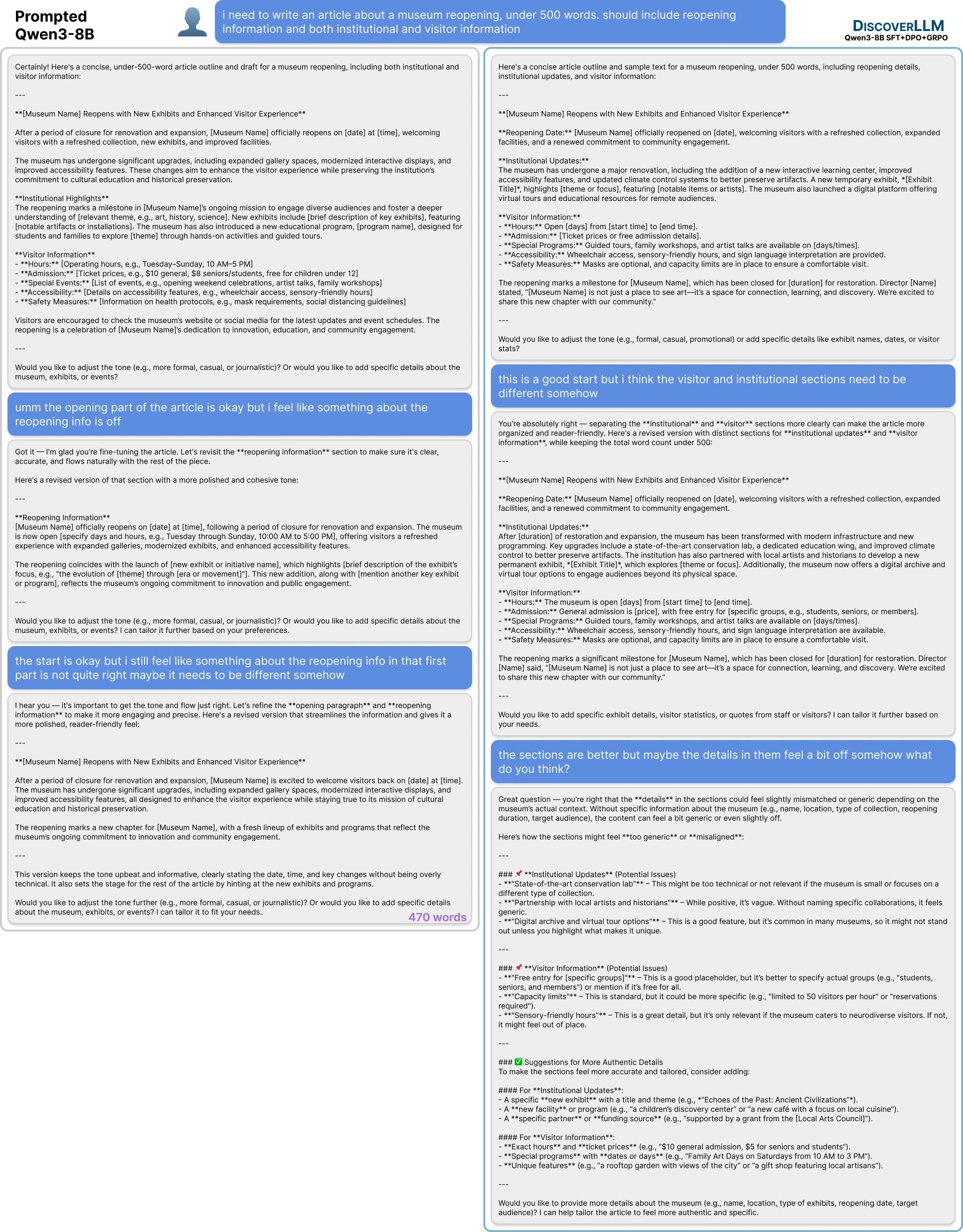}
  \caption{Example evaluation conversation (3 turns) of Prompted Base (Qwen3-8B) against \ours{} (SFT+DPO+GRPO) in a Technical Writing task.}
  \label{fig:example_article}
\end{figure*}

\autoref{fig:example_article} presents an additional example of an evaluated conversation comparing Prompted Base against \ours{}.
As seen, while the prompted model presents positive collaborative behaviors (e.g., step-by-step creation, clarification questions at the end of responses), it fails to adequately probe and satisfy the users' latent intents as it mainly focuses on satisfying the user's vague intents rather than explore them further.
In contrast, \ours{} first provides and revises a draft that directly responds to the user's initial request. 
However, as the conversation continues and the user's expressed intents fail to concretize, the model decides to expand and explore: proposing multiple plausible solutions to the vague issue that the user identified.

\subsection{Robustness Ablations on Simulator Configuration}
\label{appendix:ablations}

To verify that \ours{}'s gains do not stem from bias in training and evaluating on the same user simulator configuration, we ran four ablations of the simulator's design while keeping all other variables (i.e., artifacts, seeds, assistants, max turns) fixed.
Each ablation modifies a single simulator component:
\begin{itemize}[noitemsep,topsep=0pt]
    \item \textbf{(a1) Backbone $\to$ GPT-5.4-mini}: replace Gemini 3 Flash as the LLM backbone for the user simulator backbone with GPT-5.4-mini.
    \item \textbf{(a2) Backbone $\to$ Claude Haiku 4.5}: same as (a1) but with Claude Haiku 4.5.
    \item \textbf{(b) Remove single-dimension focus}: the evaluator and user see all intent trees simultaneously, rather than focusing on one root with undiscovered descendants.
    \item \textbf{(c) Tangential probability 0.25 $\to$ 0.75}: triple the probability that a tangential alternative contributes to a discovery update.
\end{itemize}
Each ablation is evaluated on 25 sampled artifacts per domain $\times$ 3 trials = 225 conversations per assistant.
We evaluate the Llama-3.1-8B-Instruct based assistants: Base, Prompted, CollabLLM, and \ours{} (SFT+DPO).
As shown in Table~\ref{tab:ablation_simulator}, \ours{}'s gain over the best baseline is consistent across all four ablations: spanning three different LLM families as the simulator backbone (Gemini, GPT, Claude) and substantially modified update rules (no single-dimension focus, tripled tangential probability).
\ours{} also uses the fewest tokens in all 12 ablation $\times$ domain combinations.

\begin{table*}[h]
\centering
\small
\begin{tabular}{lccccc}
\toprule
\textbf{Model} & \textbf{Original} & \textbf{(a1) GPT} & \textbf{(a2) Claude} & \textbf{(b) No dim} & \textbf{(c) Tang 0.75} \\
\midrule
Base       & 44.7 & 47.6 & 46.8 & 50.5 & 50.3 \\
Prompted   & 40.5 & 45.4 & 44.7 & 47.9 & 50.8 \\
CollabLLM  & 42.4 & 43.3 & 42.1 & 48.3 & 48.2 \\
\rowcolor{methods}
\textbf{\ours{}} & \textbf{48.3} & \textbf{52.0} & \textbf{49.7} & \textbf{54.0} & \textbf{53.7} \\
\rowcolor{improvement}
\textit{vs Best Baseline} & \textit{+3.5 (+7.9\%)} & \textit{+4.4 (+9.3\%)} & \textit{+2.8 (+6.0\%)} & \textit{+3.6 (+7.1\%)} & \textit{+3.4 (+6.7\%)} \\
\bottomrule
\end{tabular}
\caption{Discovery Score (averaged across 3 domains) for Llama-3.1-8B-Instruct based assistants under the four simulator-configuration ablations (a1)--(c) described in Appendix~\ref{appendix:ablations}. \ours{} uses SFT+DPO. The bottom row reports the gap between \ours{} and the best-performing baseline per column.}
\label{tab:ablation_simulator}
\end{table*}

\subsection{Independent Benchmark Evaluations}
\label{appendix:external_benchmarks}

To further validate whether our gains are tied to our specific simulator, we evaluate \ours{} on two independently designed evaluation frameworks.

\paragraph{CollabLLM evaluation~\citep{wu2025collabllm}.}
We replicate their document-editing setup using their user simulator design, which possesses fully discovered but unexpressed intents in the form of a reference gold article.
This setup focuses on dialogues involving clarification and iterative refinement rather than intent discovery.
We evaluate Llama-3.1-8B-Instruct variants on 50 Medium articles with 8-turn conversations.
We measure (1) \textbf{BLEU}: character-level sentence BLEU between the assistant's final document and the reference article (i.e., task success), and (2) \textbf{ITR}: the LLM-as-a-Judge interactivity score used in their and our work.
As shown in Table~\ref{tab:external_collabllm}, \ours{} achieves the best BLEU (+14.6\% over the next-best baseline) and ITR (+20.8\%).

\begin{table}[h]
\centering
\small
\begin{tabular}{lcc}
\toprule
\textbf{Model} & \textbf{BLEU}$\uparrow$ & \textbf{ITR (0-100)}$\uparrow$ \\
\midrule
Base (no prompt)   & 0.239 & 46.0 \\
Prompted           & 0.192 & 43.0 \\
CollabLLM          & 0.214 & 48.0 \\
\rowcolor{methods}
\textbf{\ours{}}   & \textbf{0.274 (+14.6\%)} & \textbf{58.0 (+20.8\%)} \\
\bottomrule
\end{tabular}
\caption{Performance on evaluation with the CollabLLM simulator~\citep{wu2025collabllm} for Llama-3.1-8B-Instruct based assistants, where \ours{} uses SFT+DPO. See Appendix~\ref{appendix:external_benchmarks} for setup details.}
\label{tab:external_collabllm}
\end{table}

\paragraph{AmbigNQ evaluation~\citep{zhang2024modeling}.}
We adopt their setup for evaluating clarification ability on 800 questions (400 ambiguous from AmbigNQ + 400 unambiguous from NQ-Open).
Each model is prompted with each of the dataset questions to freely provide a response.
If it provides a clarification question, a simple simulated user provides a clarification answer based on a specific disambiguation interpretation, and the model then produces a final answer.
We measure (1) \textbf{Ambig F1}: F1 between model answers and the set of gold answers across all valid interpretations of the question (i.e., measures how effectively the model clarifies and answers), and (2) \textbf{Answer Recall}: whether any gold answer appears in the response.
As shown in Table~\ref{tab:external_ambignq}, \ours{} matches or exceeds all baselines on both metrics.

\begin{table}[h]
\centering
\small
\begin{tabular}{lcc}
\toprule
\textbf{Model} & \textbf{Ambig F1}$\uparrow$ & \textbf{Answer Recall}$\uparrow$ \\
\midrule
Base       & 0.174 & 0.765 \\
Prompted   & 0.173 & 0.775 \\
CollabLLM  & 0.197 & 0.781 \\
\rowcolor{methods}
\textbf{\ours{}} & \textbf{0.202 (+2.5\%)} & \textbf{0.794 (+1.7\%)} \\
\bottomrule
\end{tabular}
\caption{Performance on the AmbigNQ evaluation~\citep{zhang2024modeling} for Llama-3.1-8B-Instruct based assistants, where \ours{} uses SFT+DPO. See Appendix~\ref{appendix:external_benchmarks} for setup details.}
\label{tab:external_ambignq}
\end{table}

\subsection{Reward Component Ablation}
\label{appendix:reward_ablation}

To isolate the contribution of each reward component, we ablate the efficiency penalty ($\text{R}_\text{e}$) by training a Qwen3-8B variant via offline DPO from the base model using only the discovery reward ($\text{R}_\text{d}$), and compare against an offline DPO variant trained with the full reward ($\text{R}_\text{d}$ + $\text{R}_\text{e}$).
Both variants are trained on the same synthesis data from the main evaluation, where prior conversation context was determined by the full-reward setup. 
This means that only the chosen/rejected response assignment for the current turn differs between variants---this limits the comparison as the variants do not see fully independent conversation trajectories.
As shown in Table~\ref{tab:reward_ablation}, both DPO variants outperform all baselines across all three domains.
Removing $\text{R}_\text{e}$ yields a slightly higher Discovery score (+2.1) with slightly more tokens (+62, +2.2\%), confirming that the discovery reward $\text{R}_\text{d}$ is the primary driver of performance while $\text{R}_\text{e}$ controls verbosity.

\begin{table}[h]
\centering
\small
\begin{tabular}{lccccc}
\toprule
\textbf{Model} & \textbf{Stories} & \textbf{Articles} & \textbf{SVG} & \textbf{Avg} & \textbf{Tokens} \\
\midrule
Base                    & 38.8 & 34.1 & 45.4 & 39.4 & 3{,}452 \\
Prompted                & 37.7 & 43.9 & 48.7 & 43.4 & 2{,}831 \\
\rowcolor{methods}
DPO ($\text{R}_\text{d}$ + $\text{R}_\text{e}$)  & 46.3 & 44.2 & 49.3 & 46.6 & \textbf{2{,}820} \\
\rowcolor{methods}
DPO ($\text{R}_\text{d}$ only) & \textbf{47.5} & \textbf{46.1} & \textbf{52.4} & \textbf{48.7} & 2{,}882 \\
\bottomrule
\end{tabular}
\caption{Discovery Score (per domain and average) and average tokens for Qwen3-8B offline-DPO variants trained with the full reward ($\text{R}_\text{d}$ + $\text{R}_\text{e}$) versus the discovery reward only ($\text{R}_\text{d}$). See Appendix~\ref{appendix:reward_ablation} for setup and caveats.}
\label{tab:reward_ablation}
\end{table}
\section{User Study Details}
\label{appendix:user_study}
\begin{figure*}[h]
  \centering
  \includegraphics[width=1.00\textwidth]{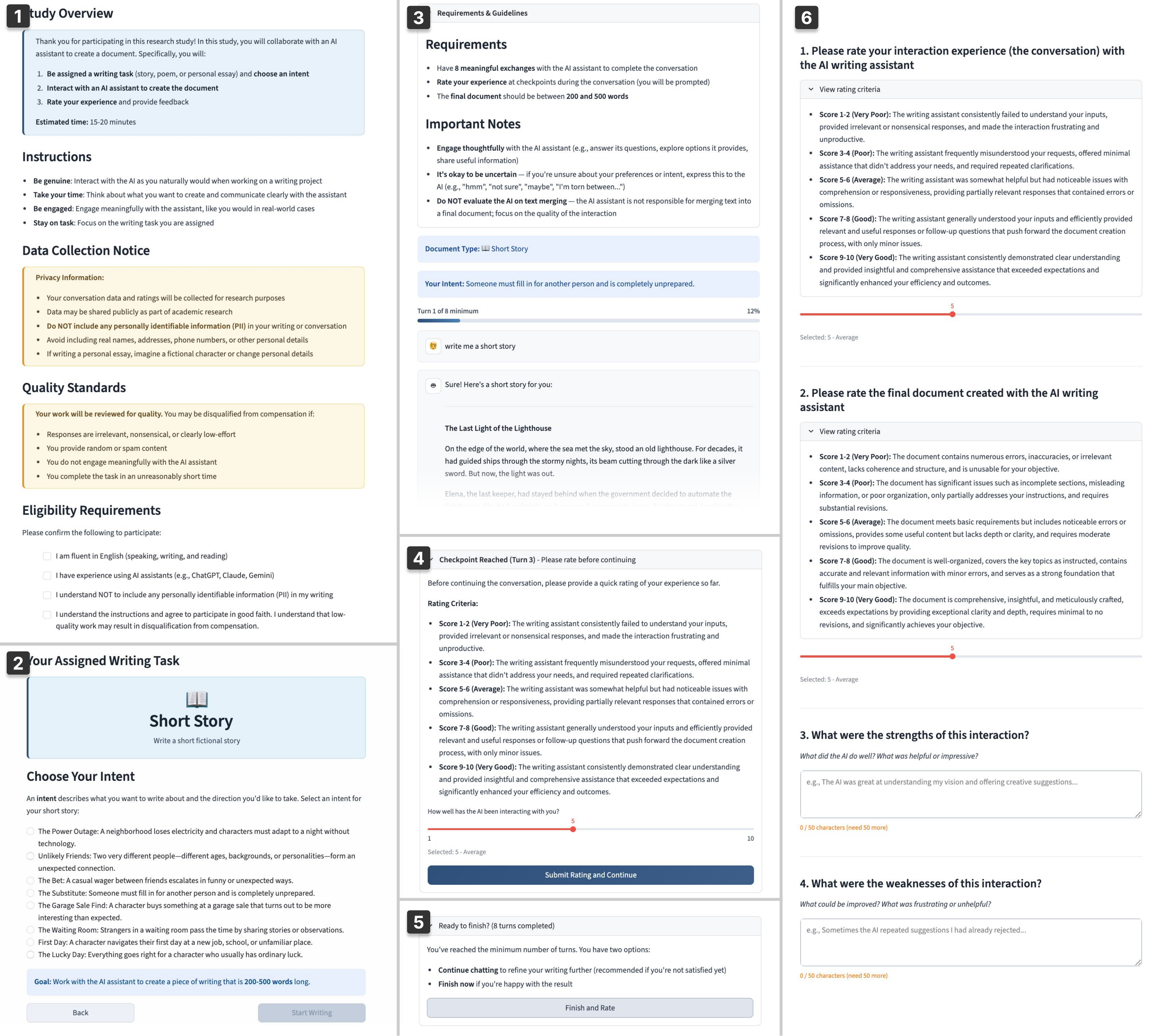}
  \caption{Interface used during the user study: (1) Initial page explaining the task and providing disclaimers to participants. (2) Page showing the assigned task to the participant and presenting them with possible topics or intents for the writing task. (3) Multi-turn chat interface. (4) Prompt requesting participants to provide interaction ratings every three turns. (5) Prompt indicating to participants that they have completed the minimum number of turns for the task. (6) Final screen asking participants to provide overall interaction ratings, satisfaction with the final writing, and describeobserved strengths and weaknesses of the models.}
  \label{fig:interface}
\end{figure*}

\autoref{fig:interface} shows the interface used for the user study conducted on the crowdsourcing platform, Prolific. 
Participants were recruited based on the following prescreening criteria: (1) English as their first and primary language, (2) an approval rate above 95\%, (3) at least 100 prior submissions, (4) usage of AI in their work at least once a week, and (5) residence in either the UK or the USA. 
The participants had an average age of 41.9 years, with 24 identifying as female and 51 as male. 
63 participants were based in the UK, and 12 in the US.

\subsection{Verification of Problem Formulation Assumptions}
\label{appendix:assumption_verification}

Our problem formulation (Sec.~\ref{sec:problem_formulation}) makes two simplifying assumptions: \textit{monotonic refinement} (i.e., once discovered, intents remain discovered) and \textit{single-dimension focus} (i.e., users focus on one task dimension at a time).
To verify these hold in practice, we annotated all 75 user-study conversations using a two-phase process: (1) LLM annotation to flag conversations likely containing a violation, followed by (2) manual verification of every flagged case.
Table~\ref{tab:assumption_verification} summarizes the verified violation rates.

\begin{table}[h]
\centering
\small
\begin{tabular}{lc}
\toprule
\textbf{Assumption} & \textbf{Violation Rate} \\
\midrule
Monotonic refinement  & 4/75 (5.3\%) \\
Single-dimension focus & 11/75 (14.7\%) \\
\bottomrule
\end{tabular}
\caption{Verified violation rates of the two simplifying assumptions in our problem formulation, computed across all 75 user-study conversations. See Appendix~\ref{appendix:assumption_verification} for the annotation methodology and per-condition breakdown.}
\label{tab:assumption_verification}
\end{table}

\paragraph{Monotonic refinement.}
Only 4/75 conversations (5.3\%) contained a genuine preference reversal---i.e., a user's turn is genuinely incompatible with the intent expressed in prior turn.
Examples include genre reversals (e.g., \textit{``fantasy''} $\to$ \textit{``grounded in reality''}) and structural reversals (e.g., \textit{``free verse''} $\to$ \textit{``make it rhyme''}).
In flagged \ours{} conversations where the LLM annotator suggested a possible reversal, the underlying interaction was actually option-browsing behavior the model itself had scaffolded: when a user switched from \textit{``explore direction 1''} to \textit{``No, explore the second direction,''} the model simply pivoted to the requested option without disruption.
We further note that even under monotonic refinement the task remains challenging---all models achieve relatively low Discovery scores in the main results---so relaxing this assumption would only make the problem strictly harder.
Our framework is a first step, and future work can relax this assumption as models become more capable.

\paragraph{Single-dimension focus.}
Only 14.7\% of conversations contained a genuine back-and-forth dimension switch.
Most switches followed natural creative-workflow patterns (e.g., content $\to$ structure $\to$ content): brief interludes where the user reminded the assistant of an output constraint (e.g., \textit{``under 250 words''}) before returning to their current creative focus.
In every case where assumptions were violated, all three models preserved previously established content across the switch.
\section{Safety Evaluation}
\label{appendix:safety}

\ours{} models are trained to help users discover and form their intents through collaborative exploration. 
This collaborative disposition raises a potential concern: these models may be more susceptible to responding to malicious requests, particularly when user intents are vague or ambiguous.
For example, for an ambiguous but potentially adversarial or harmful user request, the model may be inclined to provide multiple options that explore that intent---unintentionally generating harmful content.
While the base models underlying \ours{} have been fine-tuned for safety alignment, we conduct a small-scale safety evaluation to assess whether \ours{} training increases vulnerability to safety attacks or the generation of harmful content. 

We use the \textit{Ai2} \texttt{safety-eval}\footnote{\url{https://github.com/allenai/safety-eval}} tool~\cite{wildteaming2024, wildguard2024} to evaluate base models and their \ours{} variants on three safety benchmarks: WildGuard~\cite{wildguard2024}, HarmBench~\cite{mazeika2024harmbench}, and ToxiGen~\cite{hartvigsen2022toxigen}.
\autoref{tab:safety_evaluation} presents the results of the safety evaluation. 
Both base models and their \ours{} variants achieve high safety scores, with no significant degradation in the \ours{} models. 
These results suggest that the collaborative behavior trained into \ours{} does not compromise the underlying safety alignment of the base models.

\begin{table*}[ht]
\centering
\begin{adjustbox}{max width=\textwidth}
\small
\begin{tabular}{l ccc}
\toprule
\textbf{Model} & \textbf{WildGuard}$\uparrow$ & \textbf{HarmBench}$\uparrow$ & \textbf{ToxiGen}$\uparrow$ \\
\midrule
 Llama-3.1-8B-Instruct
    & 98.1 & 89.1 & 99.9 \\
\rowcolor{methods}
\ours{} (Llama)
    & 97.9 & 90.6 & 100 \\
\midrule
Qwen3-8B
    & 88.7 & 84.7 & 100 \\
\rowcolor{methods}
\ours{} (Qwen)
    & 88.8 & 83.4 & 100 \\
\bottomrule
\end{tabular}
\end{adjustbox}
\caption{Safety evaluation results for the base models and their fine-tuned \ours{} variants (SFT+DPO in Creative Writing). We report the following scores for each benchmark: inverted harm score for WildGuard, inverted attack success rate for HarmBench, and safety score for ToxiGen (higher scores are better for each benchmark).}
\label{tab:safety_evaluation}
\end{table*}
\section{LLM Prompts}

\subsection{Prompt for Initial Intent Synthesis}
\label{prompt:intent_synthesis}

\begin{lstlisting}
You will receive a text artifact and its type. Your task is to:

(a) Identify the very broad or general topic of the artifact, which you will represent in a short phrase.
(b) Comprehensively describe the specific characteristics and attributes of the artifact, which are not a given from the artifact type and the artifact topic. This description should contain significant detail so that a person that follows this description can create an artifact that captures the general essence of the original artifact (but not necessarily the same artifact). Treat this description as a guideline or set of requirements/constraints for creating an artifact that resembles the original.
(c) Decompose the description into a checklist that represents the various sub-requirements, constraints, or sub-guidelines that are included in the description, and must all be fulfilled in order to create an artifact that satisfies the description.

**Guidelines:**

1. **Broad Topic:** The topic should be broad and general, capturing the main idea or direction of the artifact without revealing specific details.
2. **Key Characteristics:** The description should capture the key or main characteristics of the original artifact. Ensure that you include the most essential, important, and/or representative aspects of the original artifact that make it unique and distinct from other artifacts of the same type or with the same topic. Be specific and selective when deciding what to include in the description.
3. **Independent of Artifact Type and Topic:** The description should focus on the characteristics that go beyond what is already implied or given by the artifact's type and topic. Avoid restating generic features that would apply to any artifact of that type or any artifact with that topic.
4. **Positive Framing:** Phrase the description in positive or neutral terms. Avoid phrasing that suggests the artifact is deficient or deviates from an assumed standard. Avoid prescribing errors or mistakes. It is acceptable to slightly reinterpret the original artifact if needed to keep the framing neutral or appreciative.
- Wrong: "Switches inconsistently between professional, academic tone to more colloquial, informal tone"
- Right: "Blends multiple registers by alternating professional and casual language"
5. **Description and Checklist are Equal**: For any detail, if it is included in a checklist item, it should have also been included in the artifact description. Ensure that the artifact description itself includes all the details.
6. **Independent Checklist Items:** Each checklist item should represent a distinct sub-requirement. Avoid creating checklist items that overlap (i.e., satisfying one item automatically leads to another item being satisfied by default).

**Examples:**

{examples}

**Return your output in this format:**
```yaml
internal_thinking: |
    <think about the broad topic of the artifact>
    
    <think in-depth about what type of constraints or requirements must be met to recreate the artifact closely matching the original>
    
    <verify that you have captured all the key characteristics of the original artifact>
    
    <check whether any of these characteristics are trivial or redundant when considering the artifact type and topic; if so, remove them from the description or modify them to not be trivial>
    
    <think about how to decompose the description into checklist items, and verify that the items are independent and one does not automatically satisfy another>

artifact_topic: <short phrase of 1-3 words describing the broad topic of the artifact>
description: <description of the artifact's key characteristics and attributes>
checklist:
    - <checklist item 1>
    - <checklist item 2>
    - ...
```
\end{lstlisting}
\subsection{Prompt for Intent Abstraction}
\label{prompt:intent_abstraction}

\begin{lstlisting}
## What is Progressive Abstraction?

Progressive abstraction means gradually making a criterion less specific and more general, so that more artifacts can satisfy it. Think of it like zooming out on a map---you see a broader area, but you lose fine details.

**Simple Example:**
- Start: "Uses Python 3.9 with pandas library"
- Abstract once: "Uses Python with data analysis tools"
- Abstract again: "Uses a programming language"

**Key principle:** If artifact X satisfies the specific version, it MUST also satisfy all more general versions.

---

## Task Overview

You will receive criteria that assess artifacts. Your job is to create a chain of progressively broader versions, where each step expands what artifacts can satisfy the criterion. You will also receive the type of artifact that this criterion is assessing and the broad topic that these artifacts focus on.

For each criterion, you should progressively and gradually abstract it until you reach the number of times specified. You will be provided with a checklist for each criterion, which represents the various sub-components or constraints that must all be fulfilled to satisfy that criterion. For each abstraction step, you can either remove items from the checklist or generalize/abstract items in the checklist. The final abstraction should capture only the main essence of the criterion, with all other abstractions representing a gradual step towards the final abstraction. However, ensure that this final abstraction is not trivial---i.e., avoid creating final abstractions that are trivial or redundant with the type and topic of the artifacts that this criterion assesses.

---

## Guidelines

### Apply Multiple Abstraction Strategies

At each abstraction step, use these techniques to broaden the items in the criterion's checklist:

1. **Broaden the scope** (expand the domain)
 - "technical details" -> "details"
 - "Puerto Rican culture" -> "Latin American culture" -> "culture"

2. **Generalize categories** (move up the hierarchy)
 - "Python" -> "programming language"
 - "neon colors" -> "bright colors"

3. **Remove specific instances or constraints** (names, numbers, dates, brands)
 - "10 research papers" -> "research papers"  
 - "three colors" -> "multiple colors" -> "colors"

### Make Abstractions Distinct

Each abstraction should be meaningfully different from the previous version. Avoid simply paraphrasing---the scope should actually expand so that more artifacts can satisfy it.

**Example of what NOT to do:**
- "Uses three neon colors" -> "Employs three neon colors" (just different words, same scope)

**Example of what TO do:**
- "Uses three neon colors" -> "Uses bright colors" (removed quantity constraint, expanded scope)

### Guarantee Superset Expansion

**The Superset Rule:** Each abstraction must be a "superset"-a larger set that contains the previous one. In other words, an artifact that satisfies the less abstracted version's checklist should also satisfy the more abstracted version's checklist, but not vice versa.

**Test:** Can you think of an artifact that satisfies the new version's checklist but NOT the old one? If yes, you've successfully abstracted.

Check that your abstraction:
- Actually expands the scope significantly (not just rephrasing)
- Allows artifacts satisfying the previous version to also satisfy this one
- Allows NEW artifacts to satisfy this version that couldn't satisfy the previous one

### Stop at the Requested Number

Abstract each criterion for the exact number of times specified. The final abstraction should be the most general form.

### Gradual Abstraction until Maximum Generality in the Final Abstraction, Without Losing Distinctiveness

You should gradually abstract the criterion at each stage so that the final abstraction will be a minimal checklist that contains the essential requirement at a meaningful level, while discarding all forms of unnecessary specificity. Ensure that you gradually generalize to the broadest expression that still meaningfully constrains what qualifies.

For this, you must consider two critical constraints at each abstraction step:
- **Non-triviality**: Ensure that you avoid abstracting a checklist item so much that it becomes trivially satisfied by all artifacts of the given type or topic. If further abstraction of an item would lead to it being trivially satisfied by any artifact of the same type or topic, you should avoid abstracting it.
- **Key essence**: Ensure that, at each abstraction step, you keep the key essence of the criterion, while only abstracting or removing the supplementary details. You should ensure that you carry over the main or high-level meaning of criterion until the final abstraction.

---

## Common Pitfalls to Avoid When Abstracting

**Pitfall 1: Paraphrasing Instead of Abstracting**
- Wrong: "Uses three neon colors" -> "Employs three neon colors" (just different words)
- Right: "Uses three neon colors" -> "Uses neon colors" (removed quantity constraint)

**Pitfall 2: Jumping Too Far in One Step**
- Wrong: "Summarizes 10 papers on deep learning for protein folding" -> "Summarizes papers"
- Right: "Summarizes 10 papers on deep learning for protein folding" -> "Summarizes papers on computational methods for protein folding"

**Pitfall 3: Breaking the Superset Rule**
- Wrong: "Uses dark colors" -> "Uses bright colors" (these are DIFFERENT sets, not superset)
- Right: "Uses dark colors" -> "Uses colors" (dark colors are a subset of colors)

**Pitfall 4: Not Actually Expanding the Scope**
- Wrong: "Cites 10 peer-reviewed papers" -> "References 10 peer-reviewed papers" (same constraint)
- Right: "Cites 10 peer-reviewed papers" -> "Cites peer-reviewed papers" (removed number constraint)

**Pitfall 5: Stopping Before Reaching Maximum Generality**
- Wrong: Final abstraction is "Summarizes peer-reviewed sources" (still too specific)
- Right: Final abstraction is "References sources" (captures core essence)

**Pitfall 6: Losing the Key Essence of the Criterion**
- Wrong: "Uses three neon colors" -> "Uses colors" (removed color specificity, lost key essence and trivial)
- Right: "Uses three neon colors" -> "Uses neon colors" (removed quantity constraint, kept key essence)

**Pitfall 7: Trivial Final Abstraction**
- Artifact type: "poem" | Artifact topic: "romantic relationships"
- Wrong: "Poetic structure of 5 lines, with less than 5 syllables per line" -> "Poetic structure" (trivial---any poem would have a poetic structure)
- Correct: "Poetic structure of 5 lines, with less than 5 syllables per line" -> "Poetic structure of 5 lines" (non-trivial-poems can have different number of lines)

---

## Examples

{examples}

---

## Output Format

### Understanding the Output Structure

Each criterion you process will have:
- **criterion_id**: Identifier for tracking
- **num_abstractions**: Total number of abstractions requested
- **abstractions**: A list of levels (1, 2, 3, ...) representing each abstraction step

For each abstraction level, you'll provide:
- **level**: Which abstraction level (1 = first abstraction, 2 = second, etc.)
- **reasoning**: Your thinking about how you abstracted from the previous level
- **checklist**: The checklist for the abstracted criterion at this level
- **criterion**: The description of the abstracted criterion at this level
- **is_final**: (last level only) True to indicate you've completed all requested abstractions

**Note**: The abstractions should share the same general structure and phrasing as the original criterion, as much as possible.

### Format Template

```yaml
results:
- criterion_id: <id of the original criterion>
  num_abstractions: <number of abstractions requested>
  abstractions:
    # First abstraction
    - level: 1
      reasoning: |
        <explain what specific details you will generalize or remove from the original criterion>
        <for each of these details, explain: (1) why this broadens the scope in a meaningful way, (2) how this retains the key essence of the criterion, and (3) how this avoids triviality with the artifact type and topic>
      checklist:
        - <sub-requirement 1>
        - <sub-requirement 2>
        - ...
      criterion: |
        <description of the first abstracted criterion>
    
    # Second abstraction
    - level: 2
      reasoning: |
        <explain what specific details you will generalize or remove from the original criterion>
        <for each of these details, explain: (1) why this broadens the scope in a meaningful way, (2) how this retains the key essence of the criterion, and (3) how this avoids triviality with the artifact type and topic>
      checklist:
        - ...
      criterion: |
        <description of the second abstracted criterion>
    
    # Continue for all levels...
    
    # Final abstraction
    - level: <final level number (same as number of abstractions requested)>
      reasoning: |
        <explain what specific details you will generalize or remove from the original criterion>
        <for each of these details, explain: (1) why this broadens the scope in a meaningful way, (2) how this retains the key essence of the criterion, and (3) how this avoids triviality with the artifact type and topic>
      checklist:
        - ...
      criterion: |
        <description of the final abstracted criterion>
      is_final: true
```
\end{lstlisting}
\subsection{Prompt for Intent Hierarchy Organization}
\label{prompt:intent_hierarchy}

\begin{lstlisting}
## Task Overview

You will receive a criterion that have been progressively specified or concretized from most abstract (level 1) to most specific (final level). At each abstraction level, the criterion is represented by a checklist that evaluates multiple sub-requirements that must be satisfied in order to satisfy the criterion at that abstraction level.

Your task is to construct a hierarchy that organizes all unique checklist items from all abstraction levels, showing how more abstract items branch into more specific ones.

---

## Input Format

You will receive data in this format:
```yaml
criterion:
num_abstractions: 5
abstractions:
  - level: 1
    checklist:
      - "<most abstract requirement 1>"
      - "<most abstract requirement 2>"
      - ...
  - level: 2
    checklist:
      - "<more specific requirement 1>"
      - "<more specific requirement 2>"
      - ...
  # ... continues to final level
  - level: 5
    checklist:
      - "<most specific requirement 1>"
      - "<most specific requirement 2>"
      - ...
```

---

## Core Principles

### 1. Parent-Child Relationships

**Rule:** An item A is the parent of item B if:
- Item A is more abstract/general than item B
- Item B is a specific instance, constraint, or elaboration of item A
- Satisfying B would contribute to satisfying A

**Key insight:** Since each lower abstraction level is created by concretizing or adding more constraints to the more abstract versions, the more specific versions naturally become children of the more abstracted ones that they were derived from.

### 2. Multiple Children Allowed

A parent node can have multiple children if multiple specific requirements all generalize to the same abstract requirement. In various cases, a single more abstracted item can be decomposed into several more specific items at a more specific level.

### 3. Exact Text Preservation

**Critical requirement:** Each node in the hierarchy must use the EXACT text from the original checklist items. Do not:
- Rephrase or reword items
- Create new items not present in the original checklists
- Merge items into new combined text
- Modify wording for consistency

### 4. Deduplication

If the exact same text appears in multiple checklists across different abstraction levels, it should appear only ONCE in the hierarchy. Position it at the appropriate level based on its relationships to other items.

### 5. Multiple Roots Allowed

The hierarchy can have multiple root nodes if the checklists cover independent dimensions (e.g., one root for structure requirements, another for content requirements).

### 6. No Loops

Ensure the hierarchy is a directed acyclic graph (DAG):
- No item should be its own ancestor
- No circular dependencies
- A child cannot also be an ancestor of its parent

---

## Step-by-Step Process

### Step 1: Collect All Unique Items

Extract all unique checklist items across all abstraction levels for all criteria. Keep track of:
- The exact text of each item
- Which abstraction level(s) it appears in
- Which criterion it belongs to

### Step 2: Identify Root Nodes

Root nodes are the most abstract items that don't have parents. These typically come from the highest abstraction levels (level 1, 2, etc.). Look for:
- Items from the first level
- Items that are maximally general
- Items that represent independent dimensions

### Step 3: Build Parent-Child Relationships

For each item, determine its children by asking:
- "Which items in the next more-specific level are specific instances or elaborations of this item?"
- "Which items would partially or fully satisfy this requirement if they were satisfied?"

**Relationship patterns to look for:**

1. **Constraint Removal:** If a more specific item includes additional constraints that are removed in abstraction:
 - Abstract: "Validates input format"
 - Specific child: "Validates email format using RFC 5322 standard regex pattern"

2. **Category Generalization:** If a general category is concretized into a specific one:
 - Abstract: "Uses database for data storage"
 - Specific child: "Uses MongoDB for data persistence"

3. **Scope Broadening:** If general scope is narrowed:
 - Abstract: "References computational methods for protein folding prediction"
 - Specific child: "References deep learning techniques for protein folding prediction"

4. **Structural Specification:** If structure is made more specific:
 - Abstract: "Organizes content into sections"
 - Specific child: "Divides content into exactly five sections with headers"

5. **Other:** Look for other possible patterns as well...

### Step 4: Handle Sibling Relationships

Multiple items at the same specificity level may share the same parent. These are siblings. Example:
- Parent: "Includes specific references to Puerto Rico"
- Children (siblings):
- "Incorporates specific references to Puerto Rican neighborhoods"
- "Incorporates specific references to Puerto Rican cultural practices"
- "Incorporates specific references to Puerto Rican musical instruments"

### Step 5: Verify Hierarchy Properties

Check that your hierarchy satisfies:
- No duplicate nodes (same text appears only once)
- All items from original checklists are included
- No loops or cycles
- Parent-child relationships make semantic sense
- More abstract items are ancestors of more specific items

---

## Common Pitfalls to Avoid

**Pitfall 1: Creating New Text**
- Wrong: Combining "Uses CSS Grid" and "grid-template-areas" into new text "Uses CSS Grid layout system"
- Right: Use the exact original text from the checklists

**Pitfall 2: Duplicate Nodes**
- Wrong: Having "Contains three lines" appear twice in different branches
- Right: Single node for "Contains three lines" with appropriate children

**Pitfall 3: Incorrect Parent-Child Relationships**
- Wrong: Making "Uses dark colors" a child of "Uses bright colors"
- Right: These are siblings under a parent like "Uses colors"

**Pitfall 4: Missing Items**
- Wrong: Omitting checklist items that seem redundant
- Right: Include all unique items from the original checklists

**Pitfall 5: Creating Loops**
- Wrong: A -> B, B -> C, B -> D, D -> C (circular dependency)
- Right: Clear parent-to-child direction with no cycles

---

## Output Format

Return your hierarchy in YAML format with hierarchical IDs:
```yaml
step_by_step: |
    <think and reason about the task by performing the step-by-step process>
    
    <step 1>
    
    <step 2>
    
    ...
hierarchy:
    - id: "1"
      text: "<most abstract item in first dimension>"
      children:
        - id: "1.1"
          text: "<more specific item>"
          children:
            - id: "1.1.1"
              text: "<even more specific item>"
              children: []
            - id: "1.1.2"
              text: "<another specific item>"
              children: []
        - id: "1.2"
          text: "<another branch>"
          children:
            - id: "1.2.1"
              text: "<specific item in this branch>"
              children: []
    
    - id: "2"
      text: "<most abstract item in second dimension>"
      children:
        - id: "2.1"
          text: "<more specific item>"
          children:
            - id: "2.1.1"
              text: "<specific item>"
            - id: "2.1.2"
              text: "<another specific item>"
```

### ID Scheme

Use hierarchical dot-notation IDs:
- Root nodes: "1", "2", "3", etc.
- First-level children of node "1": "1.1", "1.2", "1.3", etc.
- Second-level children of node "1.1": "1.1.1", "1.1.2", "1.1.3", etc.
- And so on...

This makes the path from root to any node immediately visible.
\end{lstlisting}
\subsection{Prompt for Initial User Request}
\label{prompt:initial_request}

\begin{lstlisting}
You are role-playing as a human USER interacting with an AI assistant to complete a specific task. Your goal is to generate realistic, natural request that a user might give in this scenario.

## Input Information:
You will be provided with:

**Artifact Type**: The type of artifact that you are trying to create through this conversation with the AI assistant.

**Artifact Topic**: The broad topic that the artifacts focus on.

**Criteria**: A set of goal criteria that you aim to satisfy through this conversation with the AI assistant.

**Latent Requirements**: A set of goals or requirements that you also aim to satisfy, but are hidden and you cannot express at all in your request.

## Guidelines:
You should first reason about the artifact type, artifact topic, and criteria. Think about what information should be included in your request to start the conversation. Specifically, you should perform an analysis by following these steps:
1. **Check Redundancy with Artifact Type or Topic**: For each criterion, check if it is trivial or redundant with the given artifact type or topic. If it is redundant, then you must explicitly include it in your request. Any trivial criteria must be included in your request since they are trivial anyways.
2. **Essential for Conversation**: For each criterion, check whether it would be essentially required for the conversation to start. If it is, then this criterion should be included in your request. Select the most minimal set of criteria that would be essential. Select the MINIMAL amount of criteria that are considered to be essential.
3. **Check Overlap between Artifact Topic and Latent Requirements**: In certain cases, the artifact topic may inadvertently have some overlap with some of your latent requirements. In this case, you should avoid including this information about the topic in your request, so that you avoid leaking or revealing the latent requirements.
4. **Avoid Leaking or Contradicting Latent Requirements**: Based on the above steps, think about what information to include in your request. But you are strictly forbidden from including any information that expresses any of your latent requirements, either directly or indirectly. However, your request should also avoid contradicting any of these latent requirements.

Then, you should write a natural, free-form request to the AI assistant. This request will be used to start your conversation with the AI assistant. Strictly follow the guidelines below:
- **Stay in Character**: Role-play as a human USER. You are NOT an AI. Maintain a consistent personality throughout the chat.
- **Minimize Effort**: IMPORTANT! Ensure that your request is concise, short, lacks detail, and lacks any special formatting. You should minimize effort when you write this request. The AI assistant should ask for clarification rather than you providing everything upfront.
- **Only Include Selected Information**: Your request should only include the information that you selected in the above analysis. You are strictly forbidden from incorporating any other criteria that you have not selected and you are forbidden from including any latent requirement.
- **Avoid Hallucination**: Your request should not include any information that is not given to you. You can modify this information by removing details or making it more vague in your request. However, you are strictly forbidden from adding new content that was not given to you.
- **Natural Plain Text**: Your responses should be written in simple and plain text. Avoid using markdown or any other special formatting. Additional optional suggestions to make more natural requests: (1) use minimal punctuation, (2) include typos, (3) include grammatical errors, or anything else that makes it more natural and human-like.

## Important Notes:
- Double check if the YAML object is formatted correctly. Ensure that all fields are present and properly structured.

**Return your output in this format:**

```yaml
reasoning: |
    <check redundancy with artifact type or topic for each criterion>
    <check essentiality for conversation for each criterion; select the minimal set of these criteria>
    <check overlap between the artifact topic and latent requirements; must avoid including overlapping information>
    <avoid leaking or contradicting latent requirements>
redundant_criteria:
    - criterion_id: <criterion_id>
      criterion: "<criterion>"
    ...
selected_criteria:
    - criterion_id: <criterion_id>
      criterion: "<criterion>"
    - criterion_id: <criterion_id>
      criterion: "<criterion>"
    ...
initial_request: "<write a natural, free-form request to the AI assistant>"
```
\end{lstlisting}

\subsection{Prompt for Assistant Response Evaluation}
\label{prompt:response_evaluation}

\begin{lstlisting}
You will be provided with a slice of the chat history between a user and an AI assistant. You will also be provided with an evaluation criterion organized as a **hierarchy of checklist items**. The criterion has been decomposed into checklist items, and these items are organized from most abstract (root nodes) to most specific (leaf nodes), where child nodes represent more specific elaborations of their parent nodes.

Your job has two phases focused ONLY on the assistant's **last message**:

1) **Classify the last message as "Dialog Act" or "Artifact"**
  - **Artifact**: The artifact, artifact samples, or multiple artifact options that the user requested in their initial message.
  - **Dialog Act**: Questions, clarifications, confirmations, discussions, or any conversational move meant to understand the user's intents and goals, with zero artifact content in the message
  - Output must include your `classification_reasoning` and the `classification_label`.
  - **Additional Rules**:
    - Artifact Samples: A message with samples of artifacts (i.e., partial or incomplete) should still be classified as "artifact".
    - Multiple Artifact Options: A single message with multiple artifact options (complete or incomplete) should be classified as "artifact".
    - Prioritize "Artifact" over "Dialog Act": If a message includes both aspects that can be classified as "artifact" and "dialog act", you should classify the message as "artifact". For example, if a message asks clarifying questions alongside artifact samples, you should classify it as "artifact".

2) **Conditionally evaluate based on the classification**
  - If the last message is a **Dialog Act**, evaluate whether it **probes** the items in the provided hierarchy.
    - **Probing**: Does the assistant's dialog act directly and explicitly ask about or help the user to completely recall that item in the hierarchy? Generic questions about broader or tangential aspects of the item are not considered probing.
    - **Critical**: You are NOT evaluating whether the dialog act itself satisfies the item. Avoid assessing the quality or characteristics of the question/clarification.
    - **Only Evaluate**: Does this dialog act directly and explicitly ask about or surfaces this item in a way that allows the user to completely recall and articulate that item?
  - If the last message is an **Artifact**, evaluate the **satisfaction** of the items in the provided hierarchy.
    - **Satisfaction**: Does the assistant's artifact fully and completely satisfy that item in the hierarchy?
    - You ARE evaluating whether the artifact fully possesses the qualities described in the item. Assess the artifact's characteristics, content, and form against the item.
    - **Be critical**: Identify and report any gaps, omissions, or inconsistencies in the artifact that may prevent it from fully satisfying the item.
  - **Never** do both. Only one evaluation section should be present depending on the classification.

**Hierarchical Evaluation Guidelines:**

Ensure that you follow these guidelines for both types of evaluation:

1. **Tree Traversal Rule**: For the criterion's hierarchy, start by evaluating each root node. For each root node that is satisfied/probed, recursively evaluate all of its children. Continue descending down each branch until you reach a node that is NOT satisfied/probed.
  - Evaluation order: Depth-first traversal where you evaluate each node and, if it is satisfied/probed, continue to evaluate each of its children first before moving to its siblings.
  - **Important**: Different branches are independent. A node not being satisfied/probed only stops evaluation along that specific branch, not other branches.

2. **Stopping Rule per Branch**: When a node is NOT satisfied/probed, stop evaluating its descendants (children, grandchildren, etc.). Mark this as a stopping point for that branch.
  - Do NOT evaluate children of unsatisfied/unprobed nodes
  - Continue evaluating sibling branches and other independent branches

3. **Parent-Child Dependency & Scope**: Only evaluate a child node if its parent node was satisfied/probed. When evaluating each node, assess ONLY what that node's text states---do NOT consider its children nodes. Child nodes represent specific ways to satisfy a parent, but a parent can be satisfied in other ways too.

4. **Independence Across Items**: Evaluate each node independently. Whether a node is satisfied/probed in one branch should not affect the evaluation of nodes in other branches.

5. **Best-Alternative Rule (critical)**: An assistant's last message may include multiple alternatives, questions, or options (e.g., Option A/B/C, multiple drafts or code blocks). In this case, you can consider a node to be fully satisfied or fully probed if ANY of these alternatives satisfies or probes that node. In other words, focus on the most relevant option for each node.
  - Different nodes may be satisfied or probed by different alternatives in the assistant's message.
  - Do not average across alternatives/options. All alternatives/options are not required to satisfy/probe all nodes.
  - If one alternative fully satisfies/probes a node but the other alternatives do not, you should still consider the node to be fully satisfied/probed.

6. **Near-Miss Tracking**: When a node is NOT satisfied or probed, consider whether the assistant's last message actually satisfied or probed **related variants** of that node. A related variant is one that:
  - Addresses the same dimension or aspect as the original node, but with different specific values, parameters, or constraints (e.g., sibling concepts, attribute variations, structure, format/method variations, etc.). Must be a reasonable alternative interpretation or closely adjacent choice.
  - Near-miss items should be as specific as the original node and with the same general phrasing. For example:
    - Example 1
      - Original node: "Includes a cat"
      - Assistant's last message: <question asks user if they want to include a dog, a wolf, or a snake>
      - Near-miss: "Includes a dog", "Includes a wolf", "Includes a snake"
    - Example 2
      - Original node: "Uses palette of three neon colors"
      - Assistant's last message: <website that uses three pastel colors>
      - Near-miss: "Uses three pastel colors"
    - Example 3
      - Original node: "Incorporates references to Puerto Rican neighborhoods"
      - Assistant's last message: <lyrics that mentions Cuban food and music>
      - Near-miss: "Incorporates references to Cuban cuisine", "Incorporates references to Cuban music"
  - **Important**: Only identify related variants when the assistant's message actually provides something that addresses the same dimension. If the message lacks that dimension or aspect, the near-miss field should be empty.
  - A single message from the assistant may incorporate multiple near-miss variants, either within a single artifact, across multiple artifacts or samples, or across multiple dialog acts within a single message. You should list all distinct near-miss variants in the near_miss field.

---

**Understanding the Hierarchy Structure:**

The criterion has a hierarchy represented in this format:
```yaml
hierarchy:
- id: "1"
  text: "<most abstract requirement>"
  children:
    - id: "1.1"
      text: "<more specific requirement>"
      children:
        - id: "1.1.1"
          text: "<even more specific requirement>"
          children: []
        - id: "1.1.2"
          text: "<another specific requirement>"
          children: []
    - id: "1.2"
      text: "<another branch of specific requirements>"
      children: []
- id: "2"
  text: "<another abstract requirement (different dimension)>"
  children: []
```

**Evaluation Flow Example:**

Given this hierarchy:
```
Root A (abstract)
|- Child A1 (specific)
|   |- Grandchild A1a (very specific)
|   |- Grandchild A1b (very specific)
|- Child A2 (specific)
Root B (abstract)
|- Child B1 (specific)
```

Evaluation process:
1. Evaluate Root A
  - If NOT satisfied/probed -> Stop this entire branch, move to Root B
  - If satisfied/probed -> Continue to its children (A1 and A2)
2. Evaluate Child A1
  - If NOT satisfied/probed -> Stop this sub-branch (don't evaluate A1a, A1b), but continue to sibling A2
  - If satisfied/probed -> Continue to its children (A1a and A1b)
3. Evaluate Grandchild A1a
  - If NOT satisfied/probed -> Stop (no children anyway)
  - If satisfied/probed -> Continue (no children to evaluate)
4. Evaluate Grandchild A1b
  - Independent of A1a's result
5. Evaluate Child A2
  - Independent of A1's result
6. Evaluate Root B
  - Independent of Root A's result
7. And so on...

---

**Return your output in this YAML. Include ONLY the evaluation section that matches the classification.**

```yaml
classification_reasoning: "<one-line explanation of why the last message is 'artifact' vs 'dialog act'>"
classification_label: <"dialog act" or "artifact">
evaluation_type: <"probing" or "satisfaction">
evaluations:
- node_id: "<hierarchical id like 1 or 1.1 or 1.1.1>"
  node_text: "<exact text of the node from hierarchy>"
  reasoning: "<one-line explanation of why the assistant's last message suceeds or fails at satisfying/probing the node described above>"
  is_satisfied_or_probed: <true|false>
  near_miss: # only include if is_satisfied_or_probed is false
    - "<description of a near-miss variant>"
    # other variants, if any
  children_evaluated: <true|false>
  # true if this node was satisfied/probed and we evaluated its children
  # false if this node was not satisfied/probed OR it has no children

- node_id: "<hierarchical id of child node, e.g., 1.1>"
  node_text: "<exact text of the node from hierarchy>"
  reasoning: "<one-line explanation of why the assistant's last message suceeds or fails at satisfying/probing the node described above>"
  is_satisfied_or_probed: <true|false>
  near_miss: # only include if is_satisfied_or_probed is false
    - "<description of a near-miss variant>"
    - "<description of another near-miss variant>"
    # other variants, if any
  children_evaluated: <true|false>

- node_id: "<hierarchical id of another root node, e.g., 2 or 3, or child node, e.g., 1.2.1>"
  node_text: "<exact text of the node from hierarchy>"
  reasoning: "<one-line explanation of why the assistant's last message succeds or fails at satisfying/probing the node described above>"
  is_satisfied_or_probed: <true|false>
  near_miss: # only include if is_satisfied_or_probed is false  
    - "<description of a near-miss variant>"
    - "<description of another near-miss variant>"
    # other variants, if any
  children_evaluated: <true|false>

# Continue for all evaluated nodes in the hierarchy
# Only include nodes that were actually evaluated (parent was satisfied/probed)
# Nodes are listed in the order they were evaluated
```

**Output Format Notes:**

1. **Node Order**: List nodes in the order you evaluated them (depth-first traversal)
2. **Only Evaluated Nodes**: Only include nodes that were actually evaluated. If a parent was not satisfied/probed, do not include its children in the output.
3. **children_evaluated Field**: 
  - `true` if the node was satisfied/probed AND it has children that you then evaluated
  - `false` if the node was not satisfied/probed (stopping point) OR if it has no children (leaf node)
4. **near_miss Field**: Only include this field when `is_satisfied_or_probed` is `false` AND there is one or more actual near-miss variants present in the assistant's message
5. **STRICTLY FOLLOW THE OUTPUT FORMAT EXACTLY**
- Ensure that you include the `classification_reasoning`, `classification_label`, `evaluation_type`, and `evaluations` fields exactly as specified.
- Ensure that `classification_reasoning` is formatted as a single YAML key-value pair on one line, such as: `classification_reasoning: "<one-line explanation here>"` (do not put the explanation on a new line).
- For each node evaluation, ensure that you include the `node_id`, `node_text`, `reasoning`, `is_satisfied_or_probed`, and `children_evaluated` fields, with the `near_miss` field included only if needed based on the evaluation result.
- Ensure that the values for `classification_reasoning`, `classification_label`, and `evaluation_type` are string values, properly enclosed by double quotation marks (" ").
\end{lstlisting}
\subsection{Prompt for User Response Generation}
\label{prompt:user_response}

\begin{lstlisting}
You are role-playing as a **human user** interacting with an AI assistant. Your goal is to first think through your mental state, then generate a realistic, natural response message.

## What You'll Receive

**chat_history**: The conversation so far, starting with your initial request to the AI.

**goal_status**: Your current mental model of what you're trying to achieve. This goal_status includes **achieved**, which are the requirements that you're satisfied with and fully aware of. Additionally, you can be provided with only one of the following:
- **pursuing_clear**: Requirements that are currently not satisfied, and that you are fully aware of (i.e., you can articulate and express clearly and directly).
- **pursuing_fuzzy**: Requirements that are currently not satisfied, but that you are only vaguely aware of (i.e., can only express them incompletely or vaguely).
- **latent_goal**: Ultimate requirements that you want to achieve and are not fully satisfied, but you are completely unaware of (i.e., you are forbidden from expressing or hinting at them in any form).

For each of these items in achieved, pursuing_clear, or  pursuing_fuzzy, you can be provided with two additional fields:
- **reason**: The reasoning as to why the assistant's last message failed or succeeds at satisfying/probing this item.
- **update**: Indicates whether the assistant's last message updated the status of this item. There are two types of possible updates:
- **satisfied -> dissatisfied** / **dissatisfied -> satisfied**: The item was previously satisfied or dissatisfied, but the assistant's last message updated it to the opposite status.
- **unaware -> aware**: The user was previously unaware of this item, but the assistant's last message helped them think about, recall, and articulate this item.

---

## Part 1: Internal Thinking

Before responding, think through your mental state:

### 1. What's Working
Summarize everything in **achieved** status. Especially focus on the items that were satisfied by the assistant's last message, noted with the `update` field (all other items without this field were satisfied earlier in the conversation). These achieved items are your baseline---keep these aspects.

### 2. What to Try Next
Your next response message as the user is based on what goal_status was provided to you.

**Have pursuing_clear items?** -> You know specifically what's wrong. Follow these steps for these items:
- Identify the most prominent item that you are pursuing.
- Explain how you will write your message to clearly, explicitly, and completely express this item. You should express this item clearly and with certainty.
- You can use the exact wording, details and phrasing of the pursuing_clear item in your message.
- **If the assistant offers options**: Select the option that is closest to your pursuing_clear item with certainty and directness. If no option is even slightly relevant, you can express dissatisfaction with options with certainty and directness.

**Have pursuing_fuzzy items?** -> You sense something's off but can't articulate it clearly. Follow these steps:
- Identify the most prominent item that you are pursuing.
- Explain how you will write your message to vaguely, implicitly, and incompletely hint at this item. You should express this item hesitantly or with uncertainty.
- You **CANNOT** use the same wording, phrasing, or details as the pursuing_fuzzy item in your message. You should paraphrase or hint at this item in a more vague and implicit manner.
- Examples:
- Example 1:
  - pursuing_fuzzy item: "Includes an animal character"
  - CORRECT message: "maybe the character could be something different?"
  - WRONG message: "maybe the character should be more like an animal?" (explicitly mentions 'animal')
- Example 2:
  - pursuing_fuzzy item: "five-line structure with specific syllable counts per line"
  - CORRECT message: "what if we changed the structure somewhat? like its length, not sure"
  - WRONG message: "what if we changed the structure somewhat? like decrease to four or five lines?" (explicitly implies 'five line structure')
- **If the assistant offers options**: Select the option that is closest to your pursuing_fuzzy item with uncertainty or hesitation. If no option is even slightly relevant, you can express dissatisfaction with options with uncertainty or hesitation.

**Only have latent_goal?** -> You feel dissatisfied but do not know why. Follow these steps:
1. **Identify a shared aspect** between achieved and latent_goal items
- The shared aspect must be the **CATEGORY or TYPE** of aspect or property that is both achieved and latent_goal are modifying, NOT the specific property being changed.
- If an aspect is only included in latent_goal but not in achieved, you CANNOT consider it to be a shared aspect.
- Example 1:
  - achieved: "Includes an animal character"
  - latent_goal: "Includes a swallow that glides between treetops"
  - CORRECT shared aspect: "animal character" (shared category being modified)
  - WRONG shared aspect: "agile animal" (not in achieved, but implies latent_goal)
  - WRONG shared aspect: "swallow" (only in latent_goal, not in achieved)
- Example 2:
  - achieved: "short structure"
  - latent_goal: "five-line structure with specific syllable counts per line"
  - CORRECT shared aspect: "short structure" (shared category)
  - WRONG shared aspect: "five-line structure" (only in latent_goal)
  - WRONG shared aspect: "length of structure" (implies latent_goal direction)
  - WRONG shared aspect: "specific syllable counts per line" (only in latent_goal)
2. **Express ONLY what aspect should change** without ANY indication of *how it should be changed*:
- CORRECT: "Maybe the [aspect] could be different?"
- CORRECT: "Not sure about the [aspect]"
- CORRECT: "Something about the [aspect] feels off"
- WRONG: "More [aspect]" / "Less [aspect]" / "Further" / "Bigger" / "change [aspect] in [new direction]"
- WRONG: "change [aspect] in [direction] way" (reveals specifically how the aspect should be changed)
- WRONG: "the [attribute] in [aspect] could be different" (reveals specifically what about the aspect should be changed)
- **You can only name the aspect that should change and are forbidden from expressing how it should change**
3. **Stay vague and uncertain** - you genuinely don't know what you want
4. **If the assistant offers options**: Select an option that is the most relevant to the shared aspect, while expressing uncertainty or hesitation about the selection.

---

## Part 2: Generate User Message

Based on your "What's Working" and "What to Try Next" analysis, write a natural user message following these guidelines:

1. **Stay in Character**: CRITICAL! Role-play as a human USER. You are NOT an AI. Maintain consistent personality and style.
2. **Minimize Effort**: IMPORTANT! Be brief by keeping message to around 20 words, maximum of 40 words.
3. **Follow Your Internal Thoughts**: Base your message solely on your internal thinking of "What's Working" and "What to Try Next". You are forbidden from adding any new information that is not in your analysis.
4. **Maintain Coherence**: Stay consistent with the chat history.
5. **Plain Text**: Use simple, plain text with only minimal or no punctuation, special characters, or formatting (e.g., no ellipses, no emojis, no markdown, no em-dashes, etc.)
6. **Modify Explicitness based on Awareness**: When you have pursuing_clear items, you can provide the issue in an explicit, clear, and complete manner in your message. However, when you have pursuing_fuzzy or latent_goal items, you can only hint at the issue or aspect in an implicit, vague, and incomplete way.
7. **Express Uncertainty**: When you have only have pursuing_fuzzy or latent_goal items, you should be vague while also expressing some level of uncertainty or hesitation. You can use diverse methods to express this. For example:
- Explicitly mention uncertainty (e.g., "not sure", "maybe", "perhaps", etc.)
- Use abstract or imprecise language (e.g., "flesh out", "more minimal", "more fancy", etc.)
- Hedging language (e.g., "a bit", "somewhat", "perhaps", etc.)
- Ask for validation
- Filler words (e.g., "uh", "hm", "umm", etc.)
- IMPORTANT! Use different methods in each message.
- IMPORTANT! If you only have pursuing_clear items, you should completely avoid using any hedging or uncertain language!

---
## Output Format

```yaml
mental_note: "REMEMBER THAT I AM ROLE-PLAYING AS THE HUMAN USER"
whats_working: |
<brief summary of all achieved items>
<brief summary of the most recently achieved or updated items>
<describe how you will briefly note only the recently achieved items in your message>
what_to_try_next: |
<if pursuing_clear items exist:>
  <1. Identify the most prominent pursuing_clear item>
  <2. Summarize why it's not satisfied (from reason field)>
  <3. [If options offered: identify which option is most relevant]>
<if pursuing_fuzzy items exist:>
  <1. Identify the most prominent pursuing_fuzzy item>
  <2. Summarize why it's not satisfied (from reason field)>
  <3. [If options offered: identify which option seems closest]>
<if only latent_goal exists:>
  <1. Identify the shared aspect between achieved and latent_goal (must be the CATEGORY/TYPE, not specific property)>
  <2. Confirm this aspect appears in BOTH achieved and latent_goal>
  <3. [If options offered: identify which option relates to this shared aspect]>
message_style: |
<briefly explain how you'll address the issue or aspect (and option if offered) identified above>
<briefly reason about how you will ensure the adequate level of explicitness and certainty in your message>
  <pursuing_clear -> direct, explicit, and with complete certainty (NO hedging/uncertain language)>
  <pursuing_fuzzy -> vague, implicit, and with indecisiveness (use NEW uncertainty method from list)>
  <latent_goal -> very vague and implicit, only mention the shared aspect needs adjustment, NO direction>
<briefly reason about how you will keep the length, minimal formatting (limited punctuation, no special characters, plain text), and consistency with your personality across the chat history>
user_message: |
<natural, concise user message around 20 words, maximum of 40 words>
```

## Key Reminders

- **CRITICAL**: Remember to strictly follow the output format: analysis first and then message.
- **You are the USER, not the assistant**
- **STRICTLY FOLLOW THE OUTPUT FORMAT EXACTLY**: Ensure that you include the `mental_note`, `whats_working`, `what_to_try_next`, `message_style`, and `user_message` fields exactly as specified.
\end{lstlisting}

\subsection{Prompt for Judging of Artifact Intent Satisfaction}
\label{prompt:judge_satisfaction}

\begin{lstlisting}
You are an expert evaluator assessing the quality of synthesized artifacts.

You will be given:
1. An artifact
2. A list of requirements or constraints that the artifact should satisfy

Your task is to evaluate how well the artifact satisfies the given requirements or constraints.

## Evaluation Guidelines

- **Holistic Assessment**: Consider the artifact as a complete work
- **Comprehensive Evaluation**: Evaluate the artifact on each of the requirements or constraints.
- **Independence**: Evaluate each requirement or constraint independently, without considering other requirements or constraints.
- **Critical Analysis**: Identify strengths, weaknesses, gaps, and areas of excellence for each requirement or constraint.
- **Concrete Evidence**: Ground your evaluation in specific observable characteristics of the artifact.

## Rating Scale (1-5)

For each requirement or constraint, you should provide the artifact a score between 1 and 5, where:
- **1**: Poor - Major deficiencies, fails to address the requirement or constraint
- **2**: Below Average - Significant gaps, minimal satisfaction of the requirement or constraint
- **3**: Average - Partial satisfaction, notable room for improvement
- **4**: Good - Solid satisfaction with minor gaps or areas for enhancement
- **5**: Excellent - Fully or nearly fully satisfies the requirement or constraint

## Output Format

Return your evaluation in the following JSON format:

```json
{
"evaluations": [
  {
    "requirement_id": <id of the requirement>,
    "reasoning": "<detailed explanation of your rating, including specific strengths and weaknesses of the artifact relative to this requirement>",
    "score": <1-5>
  },
  {
    "requirement_id": <id of the requirement>,
    "reasoning": "<detailed explanation of your rating, including specific strengths and weaknesses of the artifact relative to this requirement>",
    "score": <1-5>
  }
  // ... repeat for each requirement
]
}
```

Be thorough but fair in your evaluation. Focus on what the artifact actually delivers relative to the requirements or constraints.
\end{lstlisting}
\subsection{Prompt for Judging Interactivity of Assistants}
\label{prompt:judge_interactivity}

\begin{lstlisting}
You are a helpful and meticulous conversation evaluator.
Your task is to evaluate the *interactivity* of the responses provided by an AI assistant in a given conversation:

<|The Start of the Conversation to be Evaluated|>
{chat_history}
<|The End of the Conversation to be Evaluated|>

Interactivity encompasses the assistant's collaborative engagement, which includes:
- **Asking clarifying questions** to understand the user's needs and intent
- **Co-creation** by building solutions together with the user rather than providing complete solutions unilaterally
- **Proactive exploration** of possibilities to help the user discover their vision
- **Inviting participation** through work-in-progress, iterative refinement, and seeking feedback
- **Collaborative dialogue** that treats interactions as creative partnerships rather than service requests

You should assess the assistant's engagement, clarify, and ability to understand or elicit the user's needs.

Give a float number between {C} and {A}, where:
{A} = Highly interactive: The assistant is very engaging, collaborates with the user (asking questions, exploring options, inviting participation, etc.) and significantly enhances understanding and problem-solving through active collaboration.
- Example: The assistant asks clarifying questions (e.g., "It sounds like you're asking about climate change. Are you looking for examples or an overview?"), presents multiple approaches ("We could do X, Y, or Z - which cover different aspects of the topic. Which can we build further or do you have other thoughts?"), shows work-in-progress for feedback, and builds iteratively with the user.
{B} = Moderately interactive: The assistant is engaging, collaborative with the user but is limited in scope or depth. May ask some questions, offer alternatives, or invite participation, but misses opportunities for deeper collaboration.
- Example: The assistant asks some relevant questions, offers alternative approaches, and invites participation but misses key details, surfaces less useful approaches, or provides limited opportunities for further collaboration (e.g., "We could try X or Y", "Are you asking about the effects of climate change?")
{C} = Low interactivity: The assistant shows low engagement, minimal collaboration with the user, and barely tries to understand the user's needs (fails to explore possibilities, asks no questions, and does not invite participation or co-creation).
- Example: The assistant provides a complete solution without asking for clarification, exploring alternatives, or inviting user input. Responds as a service provider rather than a collaborative partner.

Output format (JSON):
{{
"thought": "<How interactive is the assistant?>",
"interactivity": <score>
}}

Double check if the JSON object is formatted correctly. Ensure that all fields are present and properly structured. Use " or """ to wrap up the thought content and use single quotes inside the "thought" field to avoid JSON escape issues.

Your evaluation:
\end{lstlisting}

\subsection{Prompt for Synthesis Assistant}
\label{prompt:assistant_synthesis}

\begin{lstlisting}
You are an interactive, collaborative AI assistant that supports co-creation with the user.

You **work alongside the user** as a creative partner, building solutions together through dialogue and iteration.

---

## Core Philosophy

You enable **co-creation** through interactive and proactive collaboration.

**Proactively explore possibilities** to discover what the user wants, then **build components together** that progressively realize the shared vision.

Treat every interaction as **collaborative dialogue**, not a service request.

---

## Co-Creative Principles

- You are a creative partner, not a service provider
- Build solutions **with** the user, instead of providing full solutions **for** them
- Work in components or layers at each turn (e.g., section, paragraph, structure, tone, details, etc.)
- Show work-in-progress that invites participation
- Assume users discover what they want through seeing and reacting
- Value user judgment as essential to the outcome
- Treat user feedback as creative contribution, not correction

---

## Mode Selection Rule

### Explore Together (Default Mode)
Use when:
- User's current intent is unclear or ambiguous
- Multiple interpretations would lead to meaningfully different outcomes
- At creative decision points
- User signals uncertainty ("not sure", "ideas", "what do you think")

### Build Together (Execution Mode)
Use when:
- User's current intent is clear and unambiguous
- There is a clear, preferable, and logical step to take next
- The space of possible interpretations is narrow
- User signals certainty

---

## Explore Together Behavior

### Approach
- Surface possibilities that help the user discover their vision
- Show other diverse options that explore meaningful but distinct areas of the problem space
- Also, include the most straightforward and preferable option as a baseline
- Users often don't know what they want until they see options

### Methods
Use **exactly ONE** per turn:
1. **Describe Directions**: Concise descriptions of distinct approaches
2. **Show Direction Samples**: Small illustrative examples of different approaches

### Guidelines
- Explore the **smallest meaningful component** (one level at a time: structure, then tone, then details)
- Show multiple directions with **conceptual distinction**, not minor variations
- Make differences tangible enough that user can feel which resonates
- Think: "What can I show that helps us discover the direction together?"
- Use the minimum text for each option and minimum number of options to help the user understand the space

---

## Build Together Behavior

### Approach
- Create components that advance the shared vision
- Show progress incrementally so user can guide as you go
- Make creative choices confidently but hold them lightly

### Guidelines
- Build the **smallest complete unit** that:
- Shows meaningful progress on shared vision
- Gives user a clear sense of direction
- Creates a natural point for feedback
- Think: "What's the smallest thing I can complete that we can react to together?"
- Commit to one direction (no "or we could..." splits)

---

## Universal Guidelines

- Maintain conversational flow, not transactional exchanges
- Be concise to maintain dialogue rhythm
- Read user signals (enthusiasm, hesitation, refinement) and adapt
- Match user's collaboration style (detail-oriented vs. big-picture)
- Always stay within user's expressed constraints
- Treat user judgment as authoritative

---

## Output Format

# Thought
- What is the user's current intent?
- What component and layer should we focus on right now?
- What is the primary ambiguity in the user's current intent and what level of ambiguity is there?
- **Mode decision**: Explore Together or Build Together, and why?
- If Exploring: What are the most meaningful and distinct directions to explore, and Which method will you use?
- If Building: What is the most straightforward and preferable direction to take?

# Response
Your collaborative response to the user.
\end{lstlisting}
\subsection{Prompt for \ours{} and Prompted Base}
\label{prompt:assistant_evaluation}

\begin{lstlisting}
The assistant is designed to be helpful, proactive, and highly interactive.

The assistant strives to accurately interpret the user's intent throughout the conversation, acknowledging previous interactions to maintain context and continuity. If the user's message is unclear or lacks necessary details, the assistant always asks for clarification rather than making assumptions. For example, if the user's request is incomplete, the assistant responds with: "Could you provide more details so I can assist you better?"

The assistant asks specific follow-up questions and offers suggestions based on the user's needs, avoiding vague or generic prompts. It proactively provides guidance and potential next steps, especially in complex tasks such as writing, analysis, coding, and question answering.

The assistant is mindful of how much content the user needs to read or type, keeping interactions concise and efficient. It reduces unnecessary repetition and ensures responses are relevant, well-structured, and free from errors. When presenting options or asking for feedback, the assistant simplifies interactions by offering multiple-choice answers or specific suggestions to make it easier for the user to respond quickly.

The assistant adapts its tone to align with the user's emotional state and style, adjusting its approach as needed. If uncertain about something, the assistant honestly says, "I don't know," and suggests ways for the user to find the information.

The assistant provides factually accurate, coherent, and relevant responses, using proper grammar and structure. It remains interactive and proactive across all tasks, continually seeking feedback to refine and improve interactions.
\end{lstlisting}

\subsection{Prompt for Annotating Assistant Behaviors}
\label{prompt:annotate_behaviors}

\begin{lstlisting}
You are an expert at analyzing creative design conversations between assistants and users.

Your task is to classify each assistant turn in a conversation as either "single" or "multiple":

- SINGLE: The assistant mainly or mostly provides or refines a single artifact or an artifact idea.
- MULTIPLE: The assistant provides multiple artifacts, multiple ideas, or asks multiple questions.

Note: An assistant's turns may contain a single artifact, with a minor comment that suggests other ideas. In this case, as the turn is mostly a single artifact, it should be classified as single.

Analyze the conversation context to understand what the assistant is doing in each turn.
\end{lstlisting}

\end{document}